\ifx\pdfoutput\undefined\else\pdfoutput=1\fi
\documentclass[11pt]{article}

\usepackage[utf8]{inputenc}
\usepackage[T1]{fontenc}
\usepackage{lmodern}
\usepackage{amsmath}
\usepackage{amssymb}
\usepackage{graphicx}
\usepackage{tikz}
\usepackage{pgfplots}
\usepackage{booktabs}
\usepackage{algorithm}
\usepackage{algpseudocode}
\usepackage[numbers,sort&compress]{natbib}
\usepackage{geometry}
\usepackage{microtype}
\usepackage[hidelinks]{hyperref}

\pgfplotsset{compat=1.18}
\usetikzlibrary{positioning,arrows.meta,shapes.geometric,decorations.pathreplacing,calc}

\newtheorem{definition}{Definition}[section]

\newtheorem{property}{Property}

\newcommand{\Real}{\mathbb{R}}
\newcommand{\Expect}{\mathbb{E}}
\newcommand{\Prob}{\mathbb{P}}
\newcommand{\sig}{\sigma}
\newcommand{\Agents}{\mathcal{A}}

\newcommand{\Battles}{\mathcal{B}}
\newcommand{\ind}[1]{\mathbf{1}\left[#1\right]}

\title{CoArena: Evaluating Computer-Use and Multi-Agent Systems in Real Time}

\author{
  Nitish Kovuru\\
  Coasty Research Lab, makers of CoArena.ai\\
  \texttt{nitish@coasty.ai}
  \and
  Prateek Jannu\\
  Coasty Research Lab, makers of CoArena.ai\\
  \texttt{prateek@coasty.ai}
}
\date{Preprint, September 2026}

\begin{document}
\maketitle

\begin{abstract}
Static benchmarks for computer-use agents fix a task set at release and score every system against it once. That makes them reproducible, and it lets them drift from what they should measure: a fixed task set ages, leaks into training corpora, and cannot follow how people actually use agents from week to week. CoArena measures use directly. Real users submit tasks; two systems, each a single model or a multi-agent pipeline behind the same tool interface, execute the same task concurrently in identical sandboxed desktops; users judge the two outcomes without knowing which system produced them; and a public leaderboard is refit from those judgments. The central contribution is a formal account of what makes such an evaluation real-time. We define real-time as five measurable properties, each with an equation and a worked example: continuous task arrival, live concurrent execution, online rating updates, freshness with contamination resistance, and bounded feedback latency from a failed run to a reusable training environment. The rating methodology follows in full: the Bradley-Terry pairwise model, its likelihood with weighted observations and ties, the penalized maximum-likelihood estimator, and the streaming update applied when a single vote arrives (a stochastic-gradient step on the same likelihood, recovering Elo). It gives confidence intervals from the observed information and a cluster-robust sandwich, rank bands from a parametric bootstrap, the rule by which a new system enters the board, and the convergence rate of the estimate. Vote quality is treated with inter-judge agreement statistics, redundant judging, and explicit handling of ties and abstentions. A five-system example with 211 votes is carried from the vote matrix to ratings, intervals, and rank bands. Every number is derived from stated inputs or labeled illustrative; none is a measurement of a deployed system.
\end{abstract}

\section{Introduction}
\label{sec:intro}

A computer-use agent receives an instruction in natural language and operates a desktop or a browser on a person's behalf: it reads the screen, moves a pointer, types, and delivers a result. The agent may be a single model behind a fixed scaffold, or a multi-agent system that plans, delegates and verifies internally before it acts. From the outside the two look the same, an instruction goes in and actions on a screen come out, and they can be evaluated the same way. So far that evaluation has followed the pattern set for language models: a benchmark author writes a fixed set of tasks, each with a programmatic or rubric-based checker, and every system is scored against that set \citep{xie2024osworld,zhou2024webarena,koh2024visualwebarena,deng2023mind2web}. The pattern has one clear virtue, reproducibility, and three structural weaknesses when the object of study is a system people actually use.

First, a fixed task set does not track use. The distribution of tasks people ask agents to do shifts with the software they use, the sites they visit, and the agents' own growing capabilities. A benchmark frozen in one year measures the workload of that year.

Second, a fixed task set ages into the training data of the models it is meant to measure. The concern is documented for language benchmarks \citep{sainz2023contamination,jacovi2023stop} and has motivated benchmarks that refresh their questions on a schedule \citep{white2024livebench,jain2024livecodebench}. A scheduled refresh reduces the problem; it does not remove the interval during which a released task set is both public and in use.

Third, a fixed task set is scored by a checker written in advance, and the checker encodes the author's notion of success. For open-ended computer work, whether an outcome is acceptable often depends on the person who asked. Human preference between two attempts at the same task is a different measurement, one that pairwise arenas for chatbots have collected at scale \citep{chiang2024chatbot,zheng2023judging}.

This paper describes an arena for computer-use agents built around those three observations, and it makes one claim precise: that the arena is a \emph{real-time} evaluation. The word is used loosely across evaluation platforms. We define it as a set of five measurable properties (Section~\ref{sec:formal}), each with notation, an equation where one applies, and a worked example. The properties are continuous task arrival, live concurrent execution, online rating updates, freshness with contamination resistance, and bounded feedback latency from a failed run to a reusable training environment. Continuous arrival means the evaluation has no release epochs. Live concurrent execution means that both agents run the same task at the same time in identical environments, and that execution is contemporaneous with submission. Online rating updates means that the published rating is a bounded-latency function of every admissible judgment. Freshness means that a task is executed within minutes of being written, before any training pipeline can have observed it. Bounded feedback latency means that a failure becomes something a developer can train against within a stated horizon.

The rest of the paper supplies the machinery those properties rest on. Section~\ref{sec:system} describes the arena concretely. Section~\ref{sec:rating} presents the rating methodology: the pairwise model, its likelihood, the batch and streaming estimators, intervals, rank bands, entry of new agents, and convergence. Section~\ref{sec:results} carries a small example through the whole pipeline and treats vote quality. Sections~\ref{sec:limitations} and~\ref{sec:conclusion} discuss limitations and conclude. Appendix~\ref{app:notation} collects the notation (Table~\ref{tab:notation}), Appendix~\ref{app:algorithms} lists the algorithms (Algorithms~\ref{alg:stream} to~\ref{alg:pipeline}), and Appendix~\ref{app:walkthrough} walks one battle through the interfaces.

A note on numbers. Every number in this paper is either derived from stated inputs in the text, produced by a simulation whose generative model and seed are stated, or labeled as illustrative. None is a measurement of a deployed system, and the reader should not interpret any figure here as an empirical result about a particular product.

\section{Related Work}
\label{sec:related}

\paragraph{Paired-comparison models.} The model we use for ratings originates with Zermelo's treatment of chess tournament results \citep{zermelo1929} and was developed independently by Bradley and Terry \citep{bradley1952}. Ford gave the condition under which the maximum-likelihood estimate exists: the comparison graph must be strongly connected in the sense that every agent has both a win and a loss path to every other \citep{ford1957}. Hunter analyzed minorization-maximization algorithms for the model and its generalizations \citep{hunter2004}; Newman gave an accelerated iteration \citep{newman2023}. Ties were incorporated by Rao and Kupper through a threshold parameter \citep{rao1967} and by Davidson through a separate tie propensity \citep{davidson1970}. The Elo rating system \citep{elo1978} is the sequential update familiar from chess rating; Glickman's Glicko \citep{glickman1999} and the TrueSkill system \citep{herbrich2006trueskill} add a per-player uncertainty that widens with inactivity.

\paragraph{Arenas for language models.} Chatbot Arena collected human pairwise preferences between anonymized chatbots and fit a Bradley-Terry model to them, with bootstrap confidence intervals \citep{chiang2024chatbot,zheng2023judging}. Its authors and others have examined the robustness of Elo-style ratings to vote order and sample size \citep{boubdir2023elo} and the influence of response style on human preference \citep{li2024arenahard}. Our setting differs in what is compared (a trajectory of screen actions ending in a delivered result, rather than a text response) and in the cost of a comparison (minutes of sandboxed execution rather than a single generation), which shapes the concurrency and queueing analysis of Section~\ref{sec:formal}.

\paragraph{Benchmarks for computer-use agents.} OSWorld \citep{xie2024osworld}, WebArena \citep{zhou2024webarena}, VisualWebArena \citep{koh2024visualwebarena}, Mind2Web \citep{deng2023mind2web}, and Android in the Wild \citep{rawles2023aitw} established executable environments with fixed task sets and programmatic checkers. They remain the appropriate instrument when reproducibility of a specific task is the goal. Our arena is complementary: its tasks are whatever users submit, its judgments are human, and its purpose is to track use rather than to fix a target.

\paragraph{Contamination and refreshed benchmarks.} The leakage of evaluation data into training corpora has been measured for language benchmarks \citep{sainz2023contamination}, and mitigations range from withholding test data from plain-text distribution \citep{jacovi2023stop} to periodic refresh \citep{white2024livebench,jain2024livecodebench} to adversarial, human-in-the-loop data collection \citep{kiela2021dynabench}. We treat contamination resistance as a consequence of the arrival process: a task that is written and executed within minutes cannot have been in any training set that closed before it was written (Property~\ref{prop:fresh}).

\paragraph{Uncertainty in evaluation.} Cluster-robust variance estimation \citep{white1980,liang1986} and the bootstrap \citep{efron1979} are the tools we use for intervals, following the argument that evaluation results should carry error bars that account for the dependence structure of the data \citep{miller2024errorbars}.

\section{System Design}
\label{sec:system}

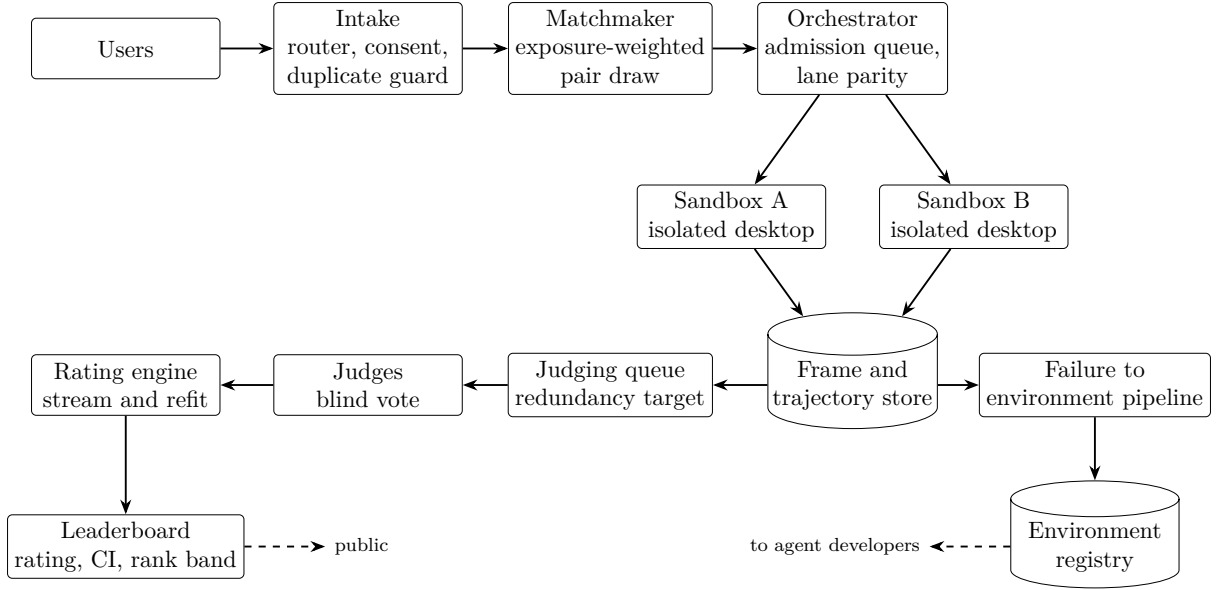
\begin{figure}[t]
\centering
\resizebox{\textwidth}{!}{%
\begin{tikzpicture}[
  font=\small,
  box/.style={draw, rounded corners=2pt, align=center, minimum height=9mm, minimum width=28mm, inner sep=3pt},
  store/.style={draw, cylinder, shape border rotate=90, aspect=0.25, align=center, minimum height=10mm, minimum width=25mm, inner sep=2pt},
  arrow/.style={-{Stealth[length=2.2mm]}, thick}
]
\node[box] (users) at (0,0) {Users};
\node[box] (intake) at (3.6,0) {Intake\\router, consent,\\duplicate guard};
\node[box] (match) at (7.2,0) {Matchmaker\\exposure-weighted\\pair draw};
\node[box] (orch) at (10.8,0) {Orchestrator\\admission queue,\\lane parity};
\node[box] (sa) at (9.0,-2.5) {Sandbox A\\isolated desktop};
\node[box] (sb) at (12.6,-2.5) {Sandbox B\\isolated desktop};
\node[box] (rating) at (0,-5.0) {Rating engine\\stream and refit};
\node[box] (judges) at (3.6,-5.0) {Judges\\blind vote};
\node[box] (queue) at (7.2,-5.0) {Judging queue\\redundancy target};
\node[store] (frames) at (10.8,-5.0) {Frame and\\trajectory store};
\node[box] (envp) at (14.4,-5.0) {Failure to\\environment pipeline};
\node[box] (board) at (0,-7.4) {Leaderboard\\rating, CI, rank band};
\node[store] (registry) at (14.4,-7.4) {Environment\\registry};

\draw[arrow] (users) to (intake);
\draw[arrow] (intake) to (match);
\draw[arrow] (match) to (orch);
\draw[arrow] (orch) to (sa);
\draw[arrow] (orch) to (sb);
\draw[arrow] (sa) to (frames);
\draw[arrow] (sb) to (frames);
\draw[arrow] (frames) to (queue);
\draw[arrow] (queue) to (judges);
\draw[arrow] (judges) to (rating);
\draw[arrow] (rating) to (board);
\draw[arrow] (frames) to (envp);
\draw[arrow] (envp) to (registry);
\draw[arrow, dashed] (board.east) to ++(1.2,0) node[right, font=\scriptsize] {public};
\draw[arrow, dashed] (registry.west) to ++(-1.2,0) node[left, font=\scriptsize] {to agent developers};
\end{tikzpicture}%
}
\caption{System architecture. A task flows left to right along the top: intake, matchmaking, and an orchestrator that provisions two isolated desktop sandboxes and starts both runs together with identical inputs. Recorded trajectories feed two consumers: the blind judging queue, whose votes drive a streaming update and a periodic refit that the leaderboard displays with intervals and rank bands; and the failure-to-environment pipeline, which converts failed runs into reproducible environments in a registry. What the figure shows is that the leaderboard and the training environments are downstream of the same recorded runs, and that nothing the submitter supplies reaches the matchmaker.}
\label{fig:architecture}
\end{figure}

This section describes the arena concretely. Figure~\ref{fig:architecture} shows the components; Figure~\ref{fig:timeline} shows one task's passage through them.

\subsection{Participants and objects}

The arena has a roster of agents $\Agents$, $|\Agents| = m$. An agent is whatever sits behind the arena's fixed tool interface and turns an instruction into actions on the desktop: a single model under the arena's own scaffold (tool set, system prompt, step budget), or a multi-agent system with its own planner, workers and verifier that the arena drives through the same interface and treats as one entrant. The arena sees only the actions and the delivered result, so a single model and a multi-agent system are rated by the same rule and can be compared on the same board. Users submit \emph{tasks}. A task $\tau$ is a natural-language instruction, optional file attachments, and an optional starting URL. A \emph{battle} $b$ pairs one task with two distinct agents $a, a' \in \Agents$ and produces two \emph{runs}, one per agent. A \emph{judgment} is a user's blind choice between the two runs of a battle.

\subsection{Intake}

Every task enters through one endpoint. The endpoint authenticates the submitter, checks that the submitter's consent to publication is current, classifies the message as a task (which will boot two sandboxes) or as conversational text that should not, applies a near-duplicate guard against the submitter's recent tasks, and admits the task to a bounded queue. Admission is deliberately wider than execution so that a burst of submissions becomes waiting time rather than an error.

\subsection{Matchmaking}

The two agents for a battle are drawn by a sampler that depends only on the roster, the count of rated battles each agent has accumulated, and the current ratings. It depends on nothing the submitter supplies. Three properties are enforced by construction and are stated here because they matter for the rating analysis. Every eligible agent has a strictly positive probability in every draw, so no agent can be starved. Agents with fewer rated battles are sampled more often, with weight proportional to $(n_a + \kappa)^{-1/2}$ for a pseudo-count $\kappa > 0$, so that measurement converges toward equal exposure. Among opponents, pairs whose outcome is uncertain under the current ratings are preferred, because those comparisons carry more information about the parameters (Section~\ref{sec:convergence}).

\subsection{Execution}

Each run executes in its own sandboxed desktop environment: an isolated virtual machine with a display, a browser, and a shell, provisioned fresh for the run. The two runs of a battle are provisioned and started together, receive byte-identical inputs (instruction, attachments, tool schemas, step budget, wall-clock budget), and execute concurrently. An agent acts through a fixed tool set (pointer, keyboard, navigation, a screenshot on every step, and a bounded shell where the environment permits it) and may end its run by delivering files and a final message. Every step's screenshot and action are recorded. Network egress from a sandbox is restricted to public destinations; private address ranges are blocked at the resolver and at the proxy.

\subsection{Judging}

When both runs have ended, the battle becomes available for judging. A judge sees the task, both trajectories as replays, and both final results, with the agents' identities withheld by the server until the judge has voted. The judge chooses one side, declares a tie, declares both unacceptable, or abstains. A random fraction of battles, drawn at creation time, is assigned a redundancy target of three judges rather than one; this sample is what agreement statistics are computed from (Section~\ref{sec:votequality}). The submitter of a task may judge their own battle, blind like anyone else; the treatment of that vote is specified in Section~\ref{sec:admissibility}.

\subsection{Rating and leaderboard}

Two estimators run. A streaming update adjusts the two agents' ratings the moment a judgment resolves a battle (Algorithm~\ref{alg:stream}). A batch estimator refits the pairwise model from every admissible judgment on a fixed period and is what the public leaderboard displays, together with confidence intervals and rank bands (Algorithms~\ref{alg:refit} and~\ref{alg:bands}). Figure~\ref{fig:convergence} shows, in a simulation, how the refit's interval narrows as judgments accumulate, and Figure~\ref{fig:arrivals} shows a simulated day of arrivals and completed runs under the capacity model of Section~\ref{sec:concurrent}. The leaderboard lists the roster only; an agent removed from the roster keeps its historical battles in the corpus, and they continue to inform the ratings of the agents it played, but it takes no position.

\subsection{Failure to environment}

A run that ends in failure (the agent gives up, the step or time budget is exhausted, or a checker judges the delivered result wrong) is a candidate for conversion into a training environment: a reproducible initial state, the instruction, and a grader derived from what went wrong (Figure~\ref{fig:pipeline}). The conversion runs as a pipeline with a stated latency target (Property~\ref{prop:feedback}).

\begin{figure}[t]
\centering
\resizebox{\textwidth}{!}{%
\begin{tikzpicture}[font=\small, x=1cm]
\draw[thick, -{Stealth[length=2.5mm]}] (0,0) to (15.8,0) node[right] {time};
\foreach \x/\lab/\sym in {0.5/submit/$s_b$, 2.6/dequeue/$q_b$, 4.6/lanes start/$q_b+\delta_b$, 8.0/lanes end/$e_b$, 11.6/judged/$h_b$, 13.4/rating updated/$r_b$, 15.0/served/$\ell_b$} {
  \draw[thick] (\x,-0.15) to (\x,0.15);
  \node[below, align=center, font=\scriptsize] at (\x,-0.2) {{\lab}\\{\sym}};
}
\draw[decorate, decoration={brace, amplitude=4pt}] (0.5,0.3) to (2.6,0.3);
\node[font=\scriptsize] at (1.55,0.72) {$L^{\mathrm{q}}_b$};
\draw[decorate, decoration={brace, amplitude=4pt}] (2.6,0.3) to (4.6,0.3);
\node[font=\scriptsize] at (3.6,0.72) {$L^{\mathrm{p}}_b$};
\draw[decorate, decoration={brace, amplitude=4pt}] (4.6,0.3) to (8.0,0.3);
\node[font=\scriptsize] at (6.3,0.72) {$L^{\mathrm{e}}_b$};
\draw[decorate, decoration={brace, amplitude=4pt}] (8.0,0.3) to (11.6,0.3);
\node[font=\scriptsize] at (9.8,0.72) {$J_b$ (human)};
\draw[decorate, decoration={brace, amplitude=4pt}] (11.6,0.3) to (13.4,0.3);
\node[font=\scriptsize] at (12.5,0.72) {$L^{\mathrm{r}}_b$};
\draw[decorate, decoration={brace, amplitude=4pt}] (13.4,0.3) to (15.0,0.3);
\node[font=\scriptsize] at (14.2,0.72) {$L^{\mathrm{s}}_b$};
\draw[decorate, decoration={brace, amplitude=5pt, mirror}] (0.5,-1.25) to (15.0,-1.25);
\node[font=\scriptsize, align=center] at (7.75,-1.85) {end-to-end latency $\ell_b - s_b$: the machine stages $L^{\mathrm{q}}_b+L^{\mathrm{p}}_b+L^{\mathrm{e}}_b+L^{\mathrm{r}}_b+L^{\mathrm{s}}_b$ are bounded by the system;\\ the human stage $J_b$ is observed, sampled, and prioritized};
\node[draw, fill=gray!10, rounded corners=2pt, align=left, font=\scriptsize] at (7.75,1.75) {Illustrative values, not measurements:\\ $L^{\mathrm{q}}=5$\,s, $L^{\mathrm{p}}=40$\,s, $L^{\mathrm{e}}=6$\,min, $J=3$\,h, $L^{\mathrm{r}}\le 31$\,s, $L^{\mathrm{s}}\le 10$\,s};
\end{tikzpicture}%
}
\caption{Timeline of one battle from submission to a served leaderboard, schematic and not to scale. The machine-controlled stages (queue wait, provisioning, execution, rating update, serving) are each bounded and add up to minutes; the human stage, waiting for a blind judge, is not under the system's control and dominates the end-to-end latency at the illustrative values shown. The point of the figure is that a real-time claim must separate the two kinds of stage: the system can guarantee bounds on its own stages and can only report, sample, and prioritize the human one.}
\label{fig:timeline}
\end{figure}
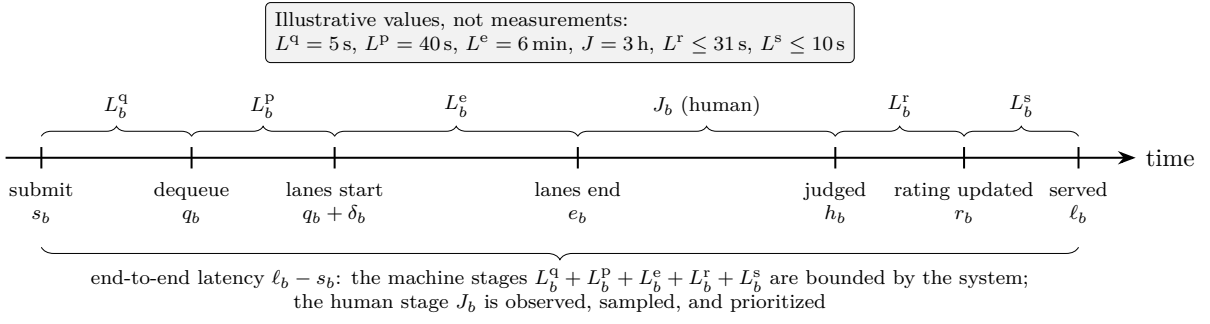

\section{Formal Model of Real-Time Evaluation}
\label{sec:formal}

This section is the paper's central contribution. We give a formal setting, then define real-time as five properties. Each property has notation, a measurable quantity, an equation where one applies, and a worked numerical example with clearly labeled illustrative inputs. We close with a composite definition and a table of time-horizon tiers.

\subsection{Setting and events}
\label{sec:setting}

Let time be $t \in \Real_{\ge 0}$. Tasks arrive as a sequence $\tau_1, \tau_2, \ldots$ with submission times $s_1 < s_2 < \cdots$. The arrival counting process is
\begin{equation}
N(t) \;=\; \#\{\, i : s_i \le t \,\}.
\label{eq:counting}
\end{equation}
Each task becomes one battle $b$ with agents $(a_b, a'_b)$. For a battle we record the event times shown in Figure~\ref{fig:timeline}: submission $s_b$; dequeue $q_b$, when the orchestrator admits the battle to execution; lane start $q_b + \delta_b$, where $\delta_b \ge 0$ is provisioning time; execution end $e_b$, when both runs have ended; first admissible judgment $h_b$; rating incorporation $r_b$, the first time a published rating is a function of that judgment; and service $\ell_b$, the first time a client is served a leaderboard that reflects it. The corresponding latencies are
\begin{equation}
L^{\mathrm{q}}_b = q_b - s_b,\quad
L^{\mathrm{p}}_b = \delta_b,\quad
L^{\mathrm{e}}_b = e_b - (q_b+\delta_b),\quad
J_b = h_b - e_b,\quad
L^{\mathrm{r}}_b = r_b - h_b,\quad
L^{\mathrm{s}}_b = \ell_b - r_b .
\label{eq:latencies}
\end{equation}
The end-to-end latency is their sum, $\ell_b - s_b$. We separate the \emph{machine latency}
\begin{equation}
M_b \;=\; L^{\mathrm{q}}_b + L^{\mathrm{p}}_b + L^{\mathrm{e}}_b + L^{\mathrm{r}}_b + L^{\mathrm{s}}_b
\label{eq:machine}
\end{equation}
from the \emph{human latency} $J_b$, because the system can bound the former and can only observe, sample, and prioritize the latter.

For a set of battles $S$ and a latency $X_b$, we write $\bar{X}(S)$ for the mean and $X_{p}(S)$ for the $p$-th percentile over $S$. Real-time properties are stated as bounds on percentiles, because a mean hides the tail that a user experiences.

\subsection{Property 1: continuous task arrival}
\label{sec:arrival}

\begin{property}[Continuous arrival]
\label{prop:arrival}
The evaluation has no release epochs. Formally, the arrival process $N(t)$ has a strictly positive long-run rate
\begin{equation}
\lambda \;=\; \lim_{t\to\infty} \frac{N(t)}{t} \;>\; 0,
\label{eq:rate}
\end{equation}
and over any observation horizon $[0,T]$ the largest gap between consecutive arrivals,
\begin{equation}
G_{\max}(T) \;=\; \max_{i:\, s_{i+1}\le T} \,(s_{i+1} - s_i),
\label{eq:gap}
\end{equation}
is small relative to $T$. A batch benchmark violates the property in the strongest way: its $N(t)$ is a step function with one jump per release, so $G_{\max}(T)$ equals the release interval.
\end{property}

The natural model for user submissions is a Poisson process \citep{kingman1993poisson}, possibly with a time-varying rate $\lambda(t)$ to capture diurnal patterns. If submissions follow a homogeneous Poisson process at rate $\lambda$, the gap between consecutive tasks has an exponential distribution with mean $1/\lambda$, and a window of length $w$ stays empty with probability
\begin{equation}
\Prob\{\text{no arrival in } w\} \;=\; e^{-\lambda w}.
\label{eq:noarrival}
\end{equation}

\paragraph{Worked example (illustrative inputs).} Take $\lambda = 12$ tasks per hour. The mean gap is $1/\lambda = 5$ minutes. The probability that a 30-minute window contains no task is $e^{-12 \cdot 0.5} = e^{-6} \approx 0.0025$. Over a 24-hour horizon the expected count is $\lambda T = 288$. Figure~\ref{fig:arrivals} shows one simulated day at this rate (280 arrivals in that draw).

\subsection{Property 2: live concurrent execution}
\label{sec:concurrent}

\begin{property}[Live concurrent execution]
\label{prop:live}
Both runs of a battle execute at the same time, in isolated environments, on identical inputs, and execution is contemporaneous with submission. Formally: (i) \emph{parity}: the resolved input tuple $I(b)$ (instruction, attachments, tool schemas, step budget $K$, wall-clock budget $W$, environment image) is identical for both lanes; (ii) \emph{co-start}: the two lane start times differ by at most a skew bound $\Delta_{\mathrm{start}}$; (iii) \emph{isolation}: the two environments share no mutable state during $[q_b, e_b]$; and (iv) \emph{contemporaneity}: the machine latency to execution end is bounded at a stated percentile,
\begin{equation}
\left(L^{\mathrm{q}} + L^{\mathrm{p}} + L^{\mathrm{e}}\right)_{p} \;\le\; H_{\mathrm{exec}} .
\label{eq:exec-bound}
\end{equation}
\end{property}

Parity is what makes a judgment a comparison of agents rather than of conditions. It is enforced by resolving every mutable input once per battle and handing the same objects to both lanes, and it is checked by a test suite rather than by review. Condition (iv) is the one that distinguishes a live arena from an evaluation that records tasks now and runs them later.

The bound in~\eqref{eq:exec-bound} is a capacity question. Let $C$ be the number of battles that may execute concurrently and let $S_b = L^{\mathrm{p}}_b + L^{\mathrm{e}}_b$ be a battle's service time, with mean $\Expect[S]$. The offered load is
\begin{equation}
\rho \;=\; \frac{\lambda\,\Expect[S]}{C},
\label{eq:utilization}
\end{equation}
and the queue is stable only if $\rho < 1$. Little's law \citep{little1961} ties the wait to the load: the expected count of battles present, whether waiting or executing, is
\begin{equation}
\Expect[\text{in system}] \;=\; \lambda \left(\Expect[L^{\mathrm{q}}] + \Expect[S]\right).
\label{eq:little}
\end{equation}

\paragraph{Worked example (illustrative inputs).} With $\lambda = 12$ per hour $= 0.2$ per minute, $\Expect[S] = 6.8$ minutes, and $C = 4$, the load is $\rho = 0.2 \times 6.8 / 4 = 0.34$. In the simulation of Figure~\ref{fig:arrivals}, which uses a lognormal service time with median 6 minutes and log-scale standard deviation $0.5$ (so $\Expect[S] = 6\,e^{0.125} = 6.80$ minutes), the mean queue wait was $0.09$ minutes and the largest wait was $2.87$ minutes; by~\eqref{eq:little} the mean number in system is $0.2 \times (0.09 + 6.80) \approx 1.4$ battles. If $\lambda$ doubled to 24 per hour at the same $C$, the load would be $\rho = 0.68$ and waits would grow sharply. The design response is to raise $C$, not to defer execution, because deferral breaks condition (iv).

\subsection{Property 3: online rating updates}
\label{sec:online}

\begin{property}[Online rating updates]
\label{prop:online}
The published rating at time $t$ is a function of every admissible judgment made before $t - L^{\mathrm{r}}$, for a bounded update latency $L^{\mathrm{r}}$ that does not grow with the size of the corpus. Define the \emph{staleness} of the published rating at time $t$ as
\begin{equation}
\varsigma(t) \;=\; t \;-\; \min\{\, h_b : h_b \le t,\; b \text{ is not yet reflected in the rating served at } t \,\},
\label{eq:staleness}
\end{equation}
with $\varsigma(t) = 0$ when no such judgment exists: the time the oldest unincorporated judgment has been waiting. A board with no pending judgment is current, however long ago the last vote arrived. The property requires $\sup_t \varsigma(t) \le H_{\mathrm{rate}}$ for a stated horizon.
\end{property}

Two estimators satisfy this property in different ways (Section~\ref{sec:rating}). The streaming update touches two parameters per judgment and has latency of milliseconds, but it depends on the order in which judgments arrived and cannot retroactively apply a retraction or a newly discovered exclusion. The batch refit is order-independent and applies every rule to the whole corpus, at the cost of a compute step. It runs on a fixed period $\Delta$ with compute time $c$, so its staleness is bounded by
\begin{equation}
\varsigma(t) \;\le\; \Delta + c .
\label{eq:staleness-bound}
\end{equation}
The arena publishes the refit as the rating and shows the streaming value beside it, labeled, because the streaming value is what a judge sees move when they vote.

\paragraph{Worked example (illustrative inputs).} With $\Delta = 30$ s and $c = 1$ s, the served rating is never more than 31 s stale. A benchmark that republishes weekly has $\varsigma$ up to $7 \times 86{,}400 = 604{,}800$ s, a ratio of about $2 \times 10^{4}$. The streaming update, with a latency of the order of $10$ ms, makes a judge's own vote visible to that judge at once; the refit makes it part of the published number within~$\Delta + c$.

\subsection{Property 4: freshness and contamination resistance}
\label{sec:fresh}

\begin{property}[Freshness and contamination resistance]
\label{prop:fresh}
A task is executed shortly after it is written, and before any agent's training pipeline could have observed it. Define the \emph{task age at execution} $A_b = q_b - s_b$ and, for agent $a$ with training-data cutoff time $t^{\mathrm{cut}}_a$, the \emph{contamination indicator}
\begin{equation}
c_b \;=\; \ind{\, s_b \le \max_{a \in \{a_b, a'_b\}} t^{\mathrm{cut}}_a \,},
\label{eq:contam}
\end{equation}
which is $1$ exactly when the task existed before the later of the two agents' cutoffs. The property requires (i) $A_{p} \le H_{\mathrm{age}}$ at a stated percentile and (ii) $c_b = 0$ for every rated battle. Because a task is written by its submitter at $s_b$ and every deployed agent's cutoff precedes deployment, (ii) holds for the arriving stream by construction; the residual risk is a \emph{repeated} task, one whose text near-duplicates an earlier public task, which is handled by the duplicate guard at intake and measured by a duplicate rate.
\end{property}

We measure freshness of the rating corpus itself by the share of rating weight that comes from tasks newer than an age $\alpha$:
\begin{equation}
F(\alpha; t) \;=\; \frac{\sum_{b:\, t - s_b \le \alpha} w_b}{\sum_{b} w_b},
\label{eq:freshness}
\end{equation}
where $w_b \in (0,1]$ is the evidence weight of battle $b$ in the likelihood (Section~\ref{sec:admissibility}) and both sums run over the battles the rating is fitted on at time $t$. Under a constant arrival rate and a corpus window of the newest $N_w$ battles with equal weights, $F(\alpha) = \min\{1, \lambda \alpha / N_w\}$.

The duplicate rate at threshold $\theta$ is
\begin{equation}
D(\theta) \;=\; \frac{1}{|\Battles|} \sum_{b} \ind{\, \max_{b' : s_{b'} < s_b} \mathrm{sim}(\tau_b, \tau_{b'}) \ge \theta \,},
\label{eq:duplicate}
\end{equation}
where $\mathrm{sim}$ is a lexical similarity (we use word-trigram Jaccard) and $\Battles$ is the rated corpus. $D(\theta)$ is a measured quantity of a deployed arena; in this paper we write it as a placeholder $D(0.55) = [\text{to be measured}]$ rather than assert a value.

\paragraph{Worked example (illustrative inputs).} If the machine stages of Figure~\ref{fig:timeline} take $L^{\mathrm{q}} = 5$ s and $L^{\mathrm{p}} = 40$ s, a task is under execution 45 s after it was written; no training pipeline with a cutoff before $s_b$ can contain it, so $c_b = 0$. With $\lambda = 12$ per hour $= 288$ per day and a fitting window of $N_w = 20{,}000$ battles, the window spans $20{,}000/288 \approx 69$ days and the share of rating weight from the last 7 days is $F(7\,\mathrm{d}) = 288 \times 7 / 20{,}000 \approx 0.10$. A fixed benchmark released once has $F(\alpha) = 0$ for every $\alpha$ shorter than its age.

\subsection{Property 5: bounded feedback latency to a training environment}
\label{sec:feedback}

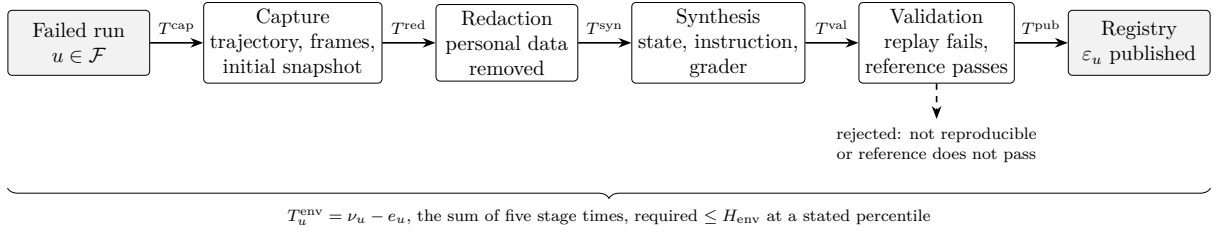
\begin{figure}[t]
\centering
\resizebox{\textwidth}{!}{%
\begin{tikzpicture}[
  font=\small,
  stage/.style={draw, rounded corners=2pt, align=center, minimum height=11mm, minimum width=24mm, inner sep=3pt},
  arrow/.style={-{Stealth[length=2.2mm]}, thick},
  node distance=9mm
]
\node[stage, fill=gray!10] (fail) {Failed run\\$u \in \mathcal{F}$};
\node[stage, right=of fail] (cap) {Capture\\trajectory, frames,\\initial snapshot};
\node[stage, right=of cap] (red) {Redaction\\personal data\\removed};
\node[stage, right=of red] (syn) {Synthesis\\state, instruction,\\grader};
\node[stage, right=of syn] (val) {Validation\\replay fails,\\reference passes};
\node[stage, right=of val, fill=gray!10] (pub) {Registry\\$\varepsilon_u$ published};
\draw[arrow] (fail) to node[above, font=\scriptsize] {$T^{\mathrm{cap}}$} (cap);
\draw[arrow] (cap) to node[above, font=\scriptsize] {$T^{\mathrm{red}}$} (red);
\draw[arrow] (red) to node[above, font=\scriptsize] {$T^{\mathrm{syn}}$} (syn);
\draw[arrow] (syn) to node[above, font=\scriptsize] {$T^{\mathrm{val}}$} (val);
\draw[arrow] (val) to node[above, font=\scriptsize] {$T^{\mathrm{pub}}$} (pub);
\draw[arrow, dashed] (val.south) to ++(0,-6mm) node[below, font=\scriptsize, align=center] {rejected: not reproducible\\or reference does not pass};
\draw[decorate, decoration={brace, amplitude=5pt, mirror}] ([yshift=-19mm]fail.south west) to ([yshift=-19mm]pub.south east);
\node[font=\scriptsize] at ($([yshift=-24mm]fail.south west)!0.5!([yshift=-24mm]pub.south east)$) {$T^{\mathrm{env}}_u = \nu_u - e_u$, the sum of five stage times, required $\le H_{\mathrm{env}}$ at a stated percentile};
\end{tikzpicture}%
}
\caption{Failure-to-environment pipeline. A failed run is captured with its trajectory and initial snapshot, redacted, synthesized into an environment triple (initial state, instruction, grader), validated by replaying the failing trajectory and a reference, and published to a registry. Three things follow from the figure: the feedback latency is a sum of five bounded stages, that validation is the dominant and run-length-dependent term, and that an environment which cannot reproduce its own failure is rejected rather than published.}
\label{fig:pipeline}
\end{figure}

\begin{property}[Bounded feedback latency]
\label{prop:feedback}
A failed run becomes a reusable, validated training environment within a stated horizon. Let $\mathcal{F}$ be the set of runs whose outcome is a failure (the agent gave up, exhausted its budget, or delivered a result a checker rejects). For $u \in \mathcal{F}$ with end time $e_u$, let $\varepsilon_u$ be the derived environment and $\nu_u$ the time it is published, validated, to the registry. The feedback latency is
\begin{equation}
T^{\mathrm{env}}_u \;=\; \nu_u - e_u \;=\; T^{\mathrm{cap}}_u + T^{\mathrm{red}}_u + T^{\mathrm{syn}}_u + T^{\mathrm{val}}_u + T^{\mathrm{pub}}_u ,
\label{eq:tenv}
\end{equation}
the sum of capture, redaction, synthesis, validation, and publication stage times (Figure~\ref{fig:pipeline}; Algorithm~\ref{alg:pipeline}). The property requires $T^{\mathrm{env}}_{p} \le H_{\mathrm{env}}$ at a stated percentile, and it requires \emph{reusability}: $\varepsilon_u$ must be reproducible from a stored initial state and a stored instruction, and \emph{validity}: replaying the failing trajectory in $\varepsilon_u$ must reproduce the failure, and a reference trajectory must pass the derived grader.
\end{property}

An environment is a triple $\varepsilon = (\text{initial state}, \text{instruction}, \text{grader})$. The initial state is a snapshot of the sandbox before the first action, the instruction is the task text after redaction of personal data, and the grader is derived from the failure mode: a budget-exhaustion failure yields a grader on the delivered artifact; a wrong-result failure yields a grader from the judged-correct run's final state where one exists. Validation runs the pipeline's own replay twice, once with the failing trajectory and once with a reference, so $T^{\mathrm{val}}$ is at least two run durations.

\paragraph{Worked example (illustrative inputs).} Take $T^{\mathrm{cap}} = 20$ s, $T^{\mathrm{red}} = 40$ s, $T^{\mathrm{syn}} = 300$ s, $T^{\mathrm{val}} = 2 \times 480 = 960$ s, and $T^{\mathrm{pub}} = 10$ s. Then $T^{\mathrm{env}} = 1{,}330$ s $\approx 22$ minutes, inside a one-hour horizon $H_{\mathrm{env}} = 3{,}600$ s. Validation dominates, and it scales with the run length, so an arena with a 20-minute wall-clock budget should expect $T^{\mathrm{env}}$ of the order of an hour at its tail.

\subsection{Composite definition and time-horizon tiers}
\label{sec:composite}

\begin{definition}[Real-time arena evaluation at horizon $H$]
\label{def:realtime}
An arena evaluation is real-time at machine horizon $H$ and environment horizon $H_{\mathrm{env}}$, at percentile $p$, if over its observation window:
\begin{enumerate}
\item (arrival) $\lambda > 0$ and $G_{\max}(T) \le H$ for every horizon $T$ of at least one day;
\item (execution) parity, co-start with $\Delta_{\mathrm{start}} \le H$, and isolation hold for every battle, and $(L^{\mathrm{q}}+L^{\mathrm{p}}+L^{\mathrm{e}})_p \le H$;
\item (rating) $\sup_t \varsigma(t) \le H$;
\item (freshness) $A_p \le H$ and $c_b = 0$ for every rated battle;
\item (feedback) $T^{\mathrm{env}}_p \le H_{\mathrm{env}}$, with reusability and validity.
\end{enumerate}
The human latency $J_b$ is reported alongside, at the same percentile, and is not part of the definition, because it is the property of a population of judges rather than of the system.
\end{definition}

Table~\ref{tab:tiers} arranges horizons into tiers. An evaluation can be real-time at one tier for some properties and not others; the honest statement names the tier per property.

\begin{table}[t]
\centering
\footnotesize
\setlength{\tabcolsep}{4pt}
\begin{tabular}{@{}llllll@{}}
\toprule
Tier & $H$ (machine) & Arrival & Execution start & Rating staleness & Task age at run \\
\midrule
T0, batch release & weeks & release epochs & offline & release interval & months to years \\
T1, daily & 24 h & daily & within the day & $\le 24$ h & $\le 24$ h \\
T2, hourly & 1 h & hourly & within the hour & $\le 1$ h & $\le 1$ h \\
T3, minutes & 10 min & gaps of minutes & within minutes & $\le 10$ min & $\le 10$ min \\
T4, seconds & 60 s & sustained stream & within seconds & $\le 60$ s & $\le 60$ s \\
\bottomrule
\end{tabular}

\vspace{4pt}
\begin{tabular}{@{}lll@{}}
\toprule
Tier & Feedback $H_{\mathrm{env}}$ & Reported human latency $J$ \\
\midrule
T0, batch release & not defined & not defined \\
T1, daily & $\le 24$ h & reported at $p$ \\
T2, hourly & $\le 1$ h & reported at $p$ \\
T3, minutes & $\le 1$ h (validation bound) & reported at $p$ \\
T4, seconds & $\le 1$ h (validation bound) & reported at $p$ \\
\bottomrule
\end{tabular}
\caption{Time-horizon tiers for the five properties of Definition~\ref{def:realtime}. Each property is assigned a tier independently; the illustrative configuration in this paper is T4 for rating staleness (a 30-second refit) and for task age at run (under a minute), T3 for execution (minutes of provisioning and a run of several minutes), and T2 for feedback (a validation stage of two run lengths). The human judging latency is reported at the same percentile and does not carry a tier.}
\label{tab:tiers}
\end{table}

\subsection{What the definition excludes}

Three designs are sometimes described as real-time and fail Definition~\ref{def:realtime}. A benchmark refreshed on a schedule satisfies freshness at the refresh interval but fails arrival (its $N(t)$ has epochs) and fails rating staleness between refreshes. A platform that records tasks continuously but executes them in nightly batches satisfies arrival and may satisfy contamination resistance, but fails execution contemporaneity, and its judgments compare runs made under conditions that may differ from the submitter's. A platform that updates a streaming rating but publishes a snapshot blended with a fixed benchmark score fails rating staleness for the published number, because the blended component moves only when the snapshot is rebuilt.

\section{Rating Methodology}
\label{sec:rating}

\subsection{Pairwise comparison model}

Each agent $a$ has a latent strength $\theta_a \in \Real$; write $\theta = (\theta_1,\ldots,\theta_m)$. Under the Bradley-Terry model \citep{bradley1952,zermelo1929} the probability that agent $i$ is preferred to agent $j$ in a comparison is
\begin{equation}
p_{ij} \;=\; \Prob\{ i \succ j \} \;=\; \frac{e^{\theta_i}}{e^{\theta_i} + e^{\theta_j}} \;=\; \sig(\theta_i - \theta_j), \qquad \sig(x) = \frac{1}{1+e^{-x}} .
\label{eq:bt}
\end{equation}
Only differences of strengths are identified; adding a constant to every $\theta_a$ leaves every $p_{ij}$ unchanged. Ratings are reported on the Elo scale by the affine map
\begin{equation}
R_a \;=\; R_0 + \frac{400}{\ln 10}\,\theta_a \;\approx\; R_0 + 173.72\,\theta_a ,
\label{eq:eloscale}
\end{equation}
with origin $R_0 = 1000$, so that a difference of 400 Elo points corresponds to odds of 10 to 1, as in \citep{elo1978}.

\subsection{Observations, ties, and weights}
\label{sec:observations}

A rated battle $b$ between agents $i_b$ and $j_b$ yields an outcome $y_b \in \{1, \tfrac{1}{2}, 0\}$: $1$ if $i_b$ was preferred, $0$ if $j_b$ was, and $\tfrac{1}{2}$ for a tie. We treat a tie as half a win for each side, the standard approximation that keeps the likelihood in the binomial family; the alternative of a tie parameter \citep{rao1967,davidson1970} adds one parameter and is appropriate when ties are frequent enough to estimate it. A judgment of ``both unacceptable'' is recorded as a distinct label, because it is a statement about quality, and enters the rating as a tie, because it carries no preference between the two.

Each battle carries an evidence weight $w_b \in (0, 1]$ (Section~\ref{sec:admissibility}), and the log-likelihood of the parameters given $n$ rated battles is
\begin{equation}
\mathcal{L}(\theta) \;=\; \sum_{b=1}^{n} w_b \Big[\, y_b \log \sig(\theta_{i_b} - \theta_{j_b}) + (1-y_b) \log \big(1 - \sig(\theta_{i_b} - \theta_{j_b})\big) \Big].
\label{eq:loglik}
\end{equation}
A battle judged by several people enters~\eqref{eq:loglik} once, with the consensus outcome of Section~\ref{sec:votequality}, not once per judge; otherwise a redundancy sample meant to measure noise would triple that battle's weight.

\subsection{Penalized maximum-likelihood estimator}
\label{sec:estimator}

Ford's condition \citep{ford1957} governs whether the unpenalized estimate exists at all: every agent must be joined to every other by both a chain of wins and a chain of losses. Early in an arena that fails routinely, because an agent that has not yet lost, or not yet won, drives its own estimate off to infinity. We therefore maximize a ridge-penalized objective,
\begin{equation}
\hat{\theta} \;=\; \arg\max_{\theta} \; \mathcal{L}(\theta) - \frac{r}{2}\,\|\theta\|_2^2 ,
\label{eq:penalized}
\end{equation}
with a small penalty $r > 0$ (we use $r = 10^{-3}$), which pins the shift and keeps a separated agent finite. The penalty is disclosed as part of the published method because it shrinks toward the origin.

Writing $x_b \in \Real^m$ for the design vector of battle $b$ ($+1$ at $i_b$, $-1$ at $j_b$, $0$ elsewhere), $\eta_b = x_b^{\top}\theta$ and $\mu_b = \sig(\eta_b)$, the gradient and negative Hessian of the objective are
\begin{equation}
g(\theta) \;=\; \sum_b w_b (y_b - \mu_b)\, x_b \;-\; r\,\theta, \qquad
H(\theta) \;=\; \sum_b w_b\, \mu_b (1-\mu_b)\, x_b x_b^{\top} \;+\; r I .
\label{eq:gradhess}
\end{equation}
We solve~\eqref{eq:penalized} by Newton steps, $\theta \leftarrow \theta + H(\theta)^{-1} g(\theta)$, stopping once no coordinate moves by more than a tolerance (Algorithm~\ref{alg:refit}). Strict concavity of the penalized objective guarantees convergence from the zero start. After convergence we center the estimate on the mean strength of the agents that have reached the provisional floor of Section~\ref{sec:entry}, so that a newcomer's first, poorly identified estimate does not move every incumbent's published number.

\subsection{Streaming update}
\label{sec:streaming}

When a single judgment resolves battle $b$ at time $h_b$, the streaming estimator takes one stochastic-gradient step on that battle's term of~\eqref{eq:loglik}:
\begin{equation}
\theta_{i_b} \leftarrow \theta_{i_b} + \eta_{i_b} w_b (y_b - \mu_b), \qquad
\theta_{j_b} \leftarrow \theta_{j_b} - \eta_{j_b} w_b (y_b - \mu_b),
\label{eq:sgd}
\end{equation}
with per-agent step sizes $\eta_a$. Mapping to the Elo scale by~\eqref{eq:eloscale} and setting $\eta_a = K_a \ln 10 / 400$ gives
\begin{equation}
R_{i_b} \leftarrow R_{i_b} + K_{i_b}\, w_b \big(y_b - \mu_b\big), \qquad \mu_b = \frac{1}{1 + 10^{(R_{j_b} - R_{i_b})/400}},
\label{eq:elo}
\end{equation}
which is the Elo update \citep{elo1978} with a per-side factor $K_a$. The streaming estimator is thus a first-order online method for the same model the refit solves exactly, and the two disagree only by the path dependence of the online method. We use a factor that decays with the agent's rated-battle count $n_a$,
\begin{equation}
K(n_a) \;=\; K_{\min} + (K_{\max} - K_{\min}) \frac{n_0}{n_0 + n_a},
\label{eq:kfactor}
\end{equation}
with $K_{\max} = 48$, $K_{\min} = 12$, and $n_0 = 30$, so that a new agent moves quickly and a settled agent slowly (Algorithm~\ref{alg:stream}). The streaming value is displayed beside the refit and labeled, because it is what changes when a judge votes, but the refit is the published rating, for the reasons given in Section~\ref{sec:online}.

\subsection{Admissibility and evidence weight}
\label{sec:admissibility}

Not every judged battle enters~\eqref{eq:loglik}. A battle is excluded, in this order, when: the judge-visible text named a model (the vote was not blind); one side has no run record; one side produced no trajectory (an upstream outage, not a performance); the submitter ended one side early; the two sides ran under different step budgets; or no admissible vote remains once retracted votes, votes by judges marked untrusted, and votes recorded in less time than the two results take to read are removed. Every exclusion is counted and published beside the rating.

Among admissible votes on a battle, votes by judges other than the submitter take precedence: if any independent judge voted, the submitter's own vote is dropped from that battle's consensus; if none has, the submitter's blind vote decides and the battle is counted in a published ``self-judged only'' tally. The evidence weight is
\begin{equation}
w_b \;=\; \gamma_b \cdot \bar{\omega}_b, \qquad \gamma_b \in \{\gamma_{\mathrm{cal}}, 1\}, \quad \bar{\omega}_b = \frac{1}{|V_b|} \sum_{v \in V_b} \omega_v \in [\omega_{\min}, 1],
\label{eq:weight}
\end{equation}
where $\gamma_{\mathrm{cal}} = 0.25$ discounts a battle whose pairing was a new submitter's calibration showcase (a pairing shaped by an input the matchmaker otherwise never sees), $V_b$ is the set of judges whose votes produced the consensus outcome, and $\omega_v \in [\omega_{\min}, 1]$ is judge $v$'s graded trust weight (Section~\ref{sec:votequality}), floored at $\omega_{\min} = 0.1$ so that a graded signal discounts but never silences.

\subsection{Confidence intervals}
\label{sec:intervals}

\paragraph{Model-based.} At the optimum, $H(\hat{\theta})$ from~\eqref{eq:gradhess} is the observed information. Because the likelihood does not identify the shift, the covariance of the centered parameters is obtained by projecting out the all-ones direction: with $P = I - \tfrac{1}{m} \mathbf{1}\mathbf{1}^{\top}$,
\begin{equation}
\widehat{\mathrm{Cov}}(\hat{\theta}) \;=\; P\, H(\hat{\theta})^{-1}\, P, \qquad
\mathrm{SE}(\hat{\theta}_a) = \sqrt{\widehat{\mathrm{Cov}}_{aa}}, \qquad
\mathrm{CI}_{95}(R_a) = \hat{R}_a \pm 1.96 \cdot \frac{400}{\ln 10}\,\mathrm{SE}(\hat{\theta}_a).
\label{eq:modelci}
\end{equation}
Without the projection, the shift direction contributes a variance of $1/(m r)$ to every diagonal entry, which for $r = 10^{-3}$ and $m = 5$ is $200$ in $\theta$ units: a spurious interval thousands of Elo points wide. The projection removes exactly that component.

\paragraph{Cluster-robust.} Votes are not independent draws: one judge casts many of them, and a judge's leanings are shared across their votes. We therefore publish a sandwich estimator \citep{white1980,liang1986} clustered on the judge,
\begin{equation}
\widehat{\mathrm{Cov}}_{\mathrm{CR}}(\hat{\theta}) \;=\; \frac{G}{G-1}\; H^{-1} \Big( \sum_{g=1}^{G} u_g u_g^{\top} \Big) H^{-1}, \qquad u_g = \sum_{b \in g} w_b (y_b - \mu_b)\, x_b ,
\label{eq:sandwich}
\end{equation}
where $g$ indexes the $G$ distinct judges whose votes decided battles and $u_g$ is the summed score of cluster $g$. The score vectors satisfy $\mathbf{1}^{\top} x_b = 0$, so the sandwich has no shift component and needs no projection. The estimator is identified only when $G$ exceeds the number of parameters; when it does not, no interval is published rather than a spuriously narrow one. Clustering on the judge is the correct choice for an arena in which each task is submitted once; clustering on the task would place one observation in every cluster and reproduce the naive standard errors.

\paragraph{Rank bands.} A rank is not a parameter, and its uncertainty is not a symmetric interval. We publish for each agent the 95\% band of positions it holds when every rating is drawn from its own normal approximation and the field is re-ranked, together with the share of draws in which it was first (Algorithm~\ref{alg:bands}). Draws are seeded per agent so that the same corpus yields the same band and adding an agent does not change another agent's draws. The band uses each agent's marginal variance and ignores the off-diagonal covariance; this errs toward wider bands, which is the safe direction for a published ordering.

\subsection{Entry of a new agent}
\label{sec:entry}

A new agent enters the roster with no rating. Until it has $n_0 = 30$ rated battles it is \emph{provisional}: the refit's estimate for it is withheld from the board, because a record with no loss, or no win, is separated and its estimate is determined by the penalty rather than by data; the streaming value is shown instead with an interval of the full scale, labeled as the streaming value. The matchmaker samples the newcomer with weight $(n_a + \kappa)^{-1/2}$, $\kappa = 4$, so its exposure converges to the field's. Once the floor is reached, the fitted rating, its interval, and its rank band are published. The agent is never given a seed rating that asserts it is average: its published number is whatever its record implies.

\subsection{Convergence}
\label{sec:convergence}

Consider two agents that play only each other, $n$ times, with true preference probability $p$. The information about $\theta_i - \theta_j$ per game is $p(1-p) \le \tfrac{1}{4}$, so the standard error of the estimated difference satisfies
\begin{equation}
\mathrm{SE}(\hat{\theta}_i - \hat{\theta}_j) \;=\; \frac{1}{\sqrt{n\,p(1-p)}} \;\ge\; \frac{2}{\sqrt{n}},
\label{eq:convergence}
\end{equation}
with equality at $p = \tfrac{1}{2}$. On the Elo scale the 95\% half-width is at least $1.96 \times 173.72 \times 2/\sqrt{n} \approx 681/\sqrt{n}$: about $68$ points after $100$ games and about $22$ points after $1{,}000$. This is why the matchmaker prefers pairs near parity: a comparison at $p = 0.9$ carries $0.09$ units of information against $0.25$ at $p = 0.5$, roughly a third as much for the same cost in sandbox minutes and judge attention. In a field of $m$ agents the standard error of a centered strength also depends on the connectivity of the comparison graph; Figure~\ref{fig:convergence} shows the empirical behavior in a simulation.

\section{Illustrative Results}
\label{sec:results}

All results in this section are derived from the inputs stated here. The vote matrix is illustrative, the ratings and intervals are computed from it by the method of Section~\ref{sec:rating}, and the two simulations state their generative models and seeds.

\subsection{Worked example: five agents, 211 votes}
\label{sec:worked}

Table~\ref{tab:worked} gives a vote matrix for five agents, A to E, and the ratings it implies. Each cell of the upper block records, for the row agent against the column agent, the counts (wins, losses, ties) from the row agent's point of view. Every vote is treated as a single admissible judgment by a distinct judge with weight $w_b = 1$, so the model-based interval~\eqref{eq:modelci} applies and the cluster-robust interval coincides with it. Newton's method from $\theta = 0$ converged in five iterations at tolerance $10^{-10}$ with $r = 10^{-3}$; the estimate was then centered on the mean of all five agents, since all have at least $n_0$ battles.

\begin{table}[t]
\centering
\small
\begin{tabular}{@{}lccccc@{}}
\toprule
Row vs.\ column & A & B & C & D & E \\
\midrule
A & \multicolumn{1}{c}{$\cdot$} & (12, 8, 4) & (15, 5, 2) & (14, 4, 2) & (16, 2, 2) \\
B & (8, 12, 4) & \multicolumn{1}{c}{$\cdot$} & (11, 9, 3) & (12, 6, 2) & (13, 4, 3) \\
C & (5, 15, 2) & (9, 11, 3) & \multicolumn{1}{c}{$\cdot$} & (10, 8, 4) & (12, 5, 3) \\
D & (4, 14, 2) & (6, 12, 2) & (8, 10, 4) & \multicolumn{1}{c}{$\cdot$} & (9, 7, 4) \\
E & (2, 16, 2) & (4, 13, 3) & (5, 12, 3) & (7, 9, 4) & \multicolumn{1}{c}{$\cdot$} \\
\bottomrule
\end{tabular}

\vspace{6pt}
\begin{tabular}{@{}lrrrrrrrrl@{}}
\toprule
Agent & $n_a$ & W & L & T & Score & $\hat{\theta}_a$ & $\mathrm{SE}(\hat{\theta}_a)$ & $\hat{R}_a$ (95\% CI) & Rank band ($\Prob\{\text{first}\}$) \\
\midrule
A & 86 & 57 & 19 & 10 & 0.721 & $+0.804$ & 0.196 & 1140 (1073, 1206) & 1 to 2 (0.973) \\
B & 87 & 44 & 31 & 12 & 0.575 & $+0.288$ & 0.181 & 1050 (988, 1111) & 1 to 3 (0.027) \\
C & 87 & 36 & 39 & 12 & 0.483 & $-0.045$ & 0.179 & 992 (931, 1053) & 2 to 4 (0.001) \\
D & 82 & 27 & 43 & 12 & 0.402 & $-0.340$ & 0.187 & 941 (877, 1005) & 3 to 5 (0.000) \\
E & 80 & 18 & 50 & 12 & 0.300 & $-0.707$ & 0.199 & 877 (809, 945) & 4 to 5 (0.000) \\
\bottomrule
\end{tabular}
\caption{Worked example. Top: an illustrative vote matrix for five agents; each cell is (wins, losses, ties) for the row agent against the column agent, 211 votes in total. Bottom: quantities derived from it by the method of Section~\ref{sec:rating}. $n_a$ is the agent's number of rated battles, Score is $(W + T/2)/n_a$, $\hat{\theta}_a$ is the centered penalized maximum-likelihood strength, SE is from the projected observed information~\eqref{eq:modelci}, $\hat{R}_a$ is on the Elo scale~\eqref{eq:eloscale} with its 95\% interval, and the rank band and probability of first place come from Algorithm~\ref{alg:bands} with 2,000 seeded draws. The table shows that 80 to 87 votes per agent separate the top agent from the bottom two but leave adjacent agents overlapping: A's interval overlaps B's, and A is first in 97.3\% of draws, not all of them.}
\label{tab:worked}
\end{table}

Three features of the result are worth stating. The point estimates are ordered as the scores are, which is expected when every agent has met every other with similar frequency. The intervals of adjacent agents overlap: A (1073 to 1206) and B (988 to 1111) share the range 1073 to 1111, and the bootstrap accordingly gives A the band 1 to 2 rather than a bare first place. And the ties, 29 of 211 votes, enter as half-wins and widen nothing in themselves; they carry less information than decisive votes only through $\mu_b(1-\mu_b)$ in~\eqref{eq:gradhess}, which does not depend on $y_b$. Had the same 211 votes been cast by three judges rather than 211, the cluster-robust estimator~\eqref{eq:sandwich} would have had $G = 3 < m$ clusters, and no interval would have been published.

\subsection{Convergence in a simulation}

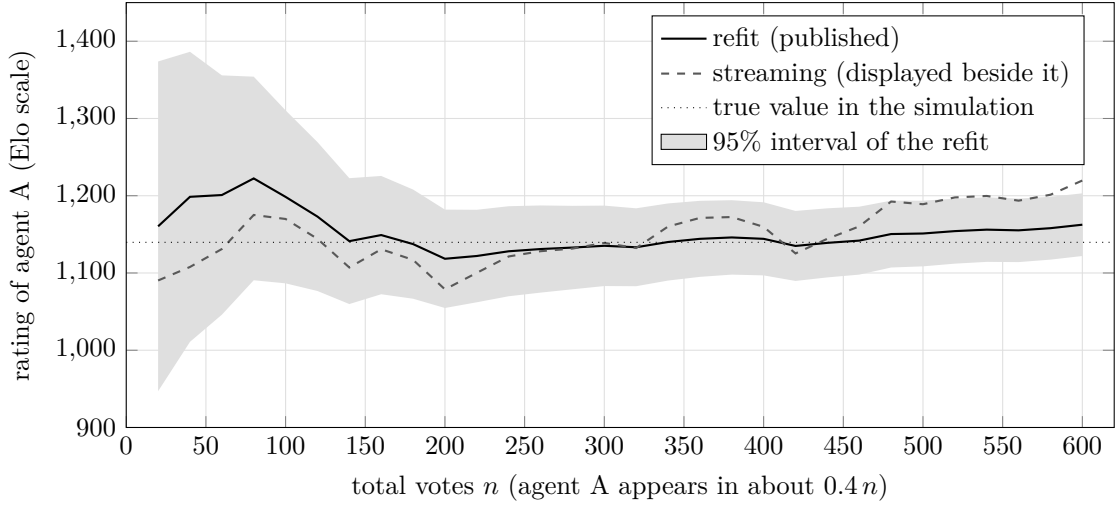
\begin{figure}[t]
\centering
\begin{tikzpicture}
\begin{axis}[
  width=0.92\textwidth, height=7.2cm,
  xlabel={total votes $n$ (agent A appears in about $0.4\,n$)},
  ylabel={rating of agent A (Elo scale)},
  xmin=0, xmax=620, ymin=900, ymax=1450,
  grid=major, grid style={gray!25},
  legend pos=north east, legend cell align=left, legend style={font=\small},
  tick label style={font=\small}, label style={font=\small}
]
\addplot[fill=gray!25, draw=none, forget plot] coordinates {
(20,1373.8) (40,1386.3) (60,1355.8) (80,1354.1) (100,1310.5) (120,1269.4) (140,1222.7) (160,1225.6) (180,1208.0) (200,1182.0) (220,1181.8) (240,1186.4) (260,1187.4) (280,1186.9) (300,1187.2) (320,1183.7) (340,1190.0) (360,1193.4) (380,1194.2) (400,1191.5) (420,1180.3) (440,1183.7) (460,1185.8) (480,1193.6) (500,1193.6) (520,1196.2) (540,1197.9) (560,1196.3) (580,1198.6) (600,1203.1) (600,1122.0) (580,1117.3) (560,1114.1) (540,1114.4) (520,1112.1) (500,1108.6) (480,1107.1) (460,1097.9) (440,1094.3) (420,1089.7) (400,1096.9) (380,1097.9) (360,1095.0) (340,1090.2) (320,1082.9) (300,1083.1) (280,1078.9) (260,1074.6) (240,1069.9) (220,1062.1) (200,1054.7) (180,1066.7) (160,1072.5) (140,1059.8) (120,1076.7) (100,1086.6) (80,1090.6) (60,1046.0) (40,1010.9) (20,947.0)
} \closedcycle;
\addplot[thick, black] coordinates {
(20,1160.4) (40,1198.6) (60,1200.9) (80,1222.4) (100,1198.5) (120,1173.0) (140,1141.2) (160,1149.1) (180,1137.4) (200,1118.4) (220,1122.0) (240,1128.1) (260,1131.0) (280,1132.9) (300,1135.2) (320,1133.3) (340,1140.1) (360,1144.2) (380,1146.0) (400,1144.2) (420,1135.0) (440,1139.0) (460,1141.9) (480,1150.4) (500,1151.1) (520,1154.2) (540,1156.1) (560,1155.2) (580,1158.0) (600,1162.5)
};
\addlegendentry{refit (published)}
\addplot[thick, dashed, gray!70!black] coordinates {
(20,1090.3) (40,1107.8) (60,1131.0) (80,1175.1) (100,1169.9) (120,1144.2) (140,1107.1) (160,1130.7) (180,1117.0) (200,1078.8) (220,1100.2) (240,1121.4) (260,1128.2) (280,1131.6) (300,1138.9) (320,1132.3) (340,1160.1) (360,1171.3) (380,1172.4) (400,1159.7) (420,1125.3) (440,1144.3) (460,1160.3) (480,1192.5) (500,1189.0) (520,1197.8) (540,1199.7) (560,1193.6) (580,1201.3) (600,1219.7)
};
\addlegendentry{streaming (displayed beside it)}
\addplot[thin, black, dotted, domain=0:620, samples=2] {1139.7};
\addlegendentry{true value in the simulation}
\addlegendimage{area legend, fill=gray!25, draw=none}
\addlegendentry{95\% interval of the refit}
\end{axis}
\end{tikzpicture}
\caption{Rating convergence for agent A in a simulation (generative model: Bradley-Terry with the strengths of Table~\ref{tab:worked}, uniform pair draws, binary outcomes, seed 7). The solid line is the penalized refit at each checkpoint of 20 votes, the shaded band is its 95\% interval from the projected observed information, the dashed line is the streaming Elo update on the same vote sequence, and the dotted line is the true value. The figure makes one point: the interval, not the point estimate, is the published claim. It it contains the truth at every checkpoint, it narrows roughly as one over the square root of the vote count (half-width 112 at 100 votes, 52 at 300, 41 at 600), and the streaming estimate is a noisier first-order approximation of the same quantity rather than a different measurement.}
\label{fig:convergence}
\end{figure}

Figure~\ref{fig:convergence} follows agent A's published rating as votes accumulate in a simulation whose generative model is the Bradley-Terry model with true strengths equal to the fitted $\hat{\theta}$ of Table~\ref{tab:worked} (so that agent A's true rating is $1139.7$ on the Elo scale), with pairs drawn uniformly from the ten possible pairs, binary outcomes drawn from~\eqref{eq:bt}, and pseudo-random seed 7. At each checkpoint of 20 total votes the refit of Section~\ref{sec:estimator} is run on the votes so far and A's rating and 95\% interval are recorded; the streaming estimator of Section~\ref{sec:streaming} runs on the same sequence from $R_0 = 1000$ with the factor~\eqref{eq:kfactor}. Agent A appears in about four of every ten votes, so 600 total votes correspond to roughly 240 of its own. The interval half-width shrinks from $112$ Elo points at $100$ total votes to $52$ at $300$ and $41$ at $600$. The interval contains the true value at every one of the 30 checkpoints, and the streaming value tracks the refit with more variability, as an online first-order method should.

\subsection{Arrivals and throughput in a simulation}

\begin{figure}[t]
\centering
\begin{tikzpicture}
\begin{axis}[
  width=0.92\textwidth, height=7cm,
  xlabel={hours since start of the simulated day},
  ylabel={cumulative count},
  xmin=0, xmax=24, ymin=0, ymax=600,
  xtick={0,4,8,12,16,20,24},
  grid=major, grid style={gray!25},
  legend pos=north west, legend cell align=left, legend style={font=\small},
  tick label style={font=\small}, label style={font=\small}
]
\addplot[const plot, thick, black] coordinates {
(0.00,0) (0.25,2) (0.50,5) (0.75,9) (1.00,14) (1.25,16) (1.50,17) (1.75,18) (2.00,25) (2.25,29) (2.50,33) (2.75,36) (3.00,36) (3.25,37) (3.50,37) (3.75,39) (4.00,44) (4.25,46) (4.50,48) (4.75,52) (5.00,54) (5.25,54) (5.50,58) (5.75,62) (6.00,63) (6.25,70) (6.50,75) (6.75,80) (7.00,81) (7.25,83) (7.50,86) (7.75,89) (8.00,89) (8.25,95) (8.50,101) (8.75,102) (9.00,105) (9.25,108) (9.50,112) (9.75,113) (10.00,115) (10.25,116) (10.50,121) (10.75,123) (11.00,126) (11.25,132) (11.50,134) (11.75,136) (12.00,144) (12.25,148) (12.50,148) (12.75,150) (13.00,155) (13.25,157) (13.50,159) (13.75,163) (14.00,163) (14.25,163) (14.50,166) (14.75,168) (15.00,170) (15.25,172) (15.50,174) (15.75,176) (16.00,178) (16.25,181) (16.50,185) (16.75,186) (17.00,187) (17.25,189) (17.50,193) (17.75,195) (18.00,200) (18.25,203) (18.50,207) (18.75,210) (19.00,213) (19.25,217) (19.50,219) (19.75,220) (20.00,225) (20.25,227) (20.50,229) (20.75,232) (21.00,238) (21.25,240) (21.50,240) (21.75,246) (22.00,249) (22.25,254) (22.50,260) (22.75,262) (23.00,263) (23.25,269) (23.50,274) (23.75,277) (24.00,280)
};
\addlegendentry{tasks arrived, $N(t)$}
\addplot[const plot, thick, dashed, gray!70!black] coordinates {
(0.00,0) (0.25,4) (0.50,4) (0.75,16) (1.00,24) (1.25,30) (1.50,32) (1.75,34) (2.00,38) (2.25,56) (2.50,64) (2.75,70) (3.00,72) (3.25,74) (3.50,74) (3.75,74) (4.00,82) (4.25,92) (4.50,96) (4.75,98) (5.00,104) (5.25,108) (5.50,116) (5.75,122) (6.00,124) (6.25,134) (6.50,144) (6.75,156) (7.00,160) (7.25,166) (7.50,172) (7.75,176) (8.00,178) (8.25,186) (8.50,198) (8.75,204) (9.00,208) (9.25,214) (9.50,218) (9.75,226) (10.00,230) (10.25,232) (10.50,238) (10.75,242) (11.00,250) (11.25,258) (11.50,266) (11.75,272) (12.00,286) (12.25,294) (12.50,296) (12.75,300) (13.00,310) (13.25,312) (13.50,314) (13.75,320) (14.00,324) (14.25,326) (14.50,330) (14.75,334) (15.00,338) (15.25,342) (15.50,344) (15.75,350) (16.00,354) (16.25,358) (16.50,366) (16.75,372) (17.00,372) (17.25,376) (17.50,380) (17.75,388) (18.00,398) (18.25,404) (18.50,412) (18.75,420) (19.00,426) (19.25,430) (19.50,436) (19.75,440) (20.00,446) (20.25,452) (20.50,458) (20.75,458) (21.00,470) (21.25,478) (21.50,480) (21.75,488) (22.00,494) (22.25,502) (22.50,516) (22.75,524) (23.00,526) (23.25,532) (23.50,542) (23.75,552) (24.00,558)
};
\addlegendentry{runs completed (two per battle)}
\addplot[thin, dotted, black, domain=0:24, samples=2] {12*x};
\addlegendentry{expected arrivals $\lambda t$, $\lambda = 12$ per hour}
\end{axis}
\end{tikzpicture}
\caption{Task arrival and cumulative completed runs over one simulated day (Poisson arrivals at 12 per hour, lognormal service with median 6 minutes, four concurrent slots, seed 11; not measured data). The solid step is the arrival count $N(t)$, the dashed step is twice the number of completed battles (each battle is two runs), and the dotted line is the expected arrival count. What to read off the figure is that under a load of $\rho = 0.34$ execution keeps pace with arrival throughout the day with a lag of about one service time, which is what Property~\ref{prop:live} requires; the small plateaus in the arrival step are the Poisson gaps whose probability equation~\eqref{eq:noarrival} gives.}
\label{fig:arrivals}
\end{figure}
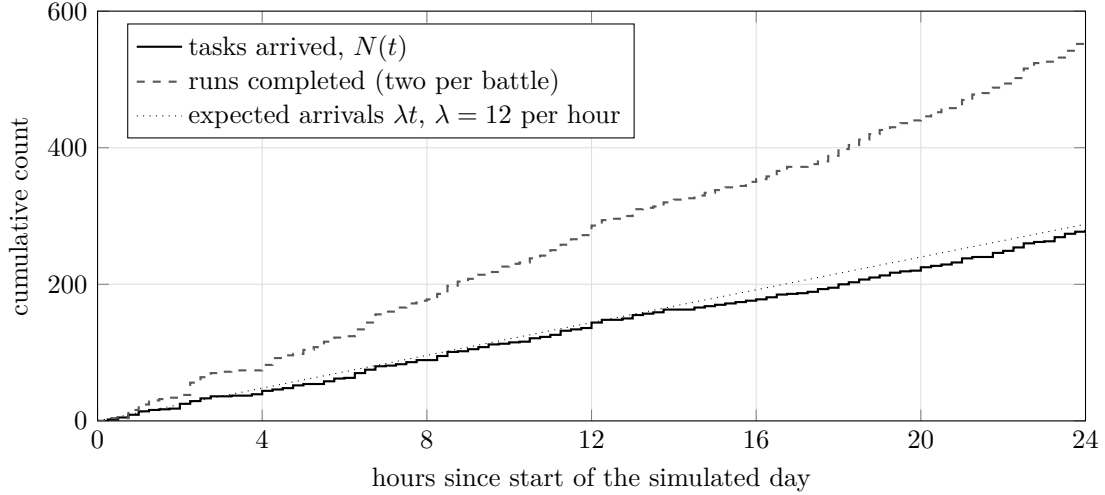

Figure~\ref{fig:arrivals} shows one simulated day of the arrival and execution model of Properties~\ref{prop:arrival} and~\ref{prop:live}: a homogeneous Poisson arrival process at $\lambda = 12$ per hour, lognormal battle service times with median 6 minutes and log-scale standard deviation $0.5$, four concurrent execution slots served first-in first-out, and seed 11. The day produced 280 arrivals, of which 279 battles (558 runs) had completed by the end of the day; the mean queue wait was $0.09$ minutes and the largest $2.87$ minutes, consistent with the load $\rho = 0.34$ computed in Section~\ref{sec:concurrent}. The cumulative-runs curve stays close to twice the arrivals curve, which is the visual signature of contemporaneous execution: work is completed at the rate it arrives, with a lag of one service time.

\subsection{Vote quality}
\label{sec:votequality}

\paragraph{Consensus on a battle.} When several judges vote on one battle, the outcome entering~\eqref{eq:loglik} is the plurality choice, with the rule that a side wins only if it out-polls both the other side and the non-preferences (ties and ``both unacceptable'') combined; otherwise the battle is a tie. One vote per person is counted, the earliest, so that a judge who voted under two identities does not carry a majority alone.

\paragraph{Abstentions.} A judge may decline to judge a battle. An abstention is recorded with its reason, releases the judge's hold so the battle returns to the queue, and does not enter the likelihood or the agreement statistics. Abstention rates are published per reason because a high rate on a class of tasks is information about the tasks.

\paragraph{Agreement.} Agreement is measured on the redundancy sample, the battles assigned three judges at creation. Drawing the sample at creation rather than at judging time is deliberate: a sample of battles that happened to attract several judges would over-represent short, recent, popular tasks. For two judges we report Cohen's $\kappa$ \citep{cohen1960}; for the three-judge sample we report Fleiss' $\kappa$ \citep{fleiss1971} and Krippendorff's $\alpha$ \citep{krippendorff2018}, the latter because it handles missing judgments and an ordinal scale over the categories (side A, tie, side B). For $N$ battles, each seen by $n$ judges choosing among categories $k$, Fleiss' $\kappa$ is
\begin{equation}
\kappa \;=\; \frac{\bar{P} - \bar{P}_e}{1 - \bar{P}_e}, \qquad
\bar{P} = \frac{1}{N} \sum_{i=1}^{N} \frac{\sum_k n_{ik}^2 - n}{n(n-1)}, \qquad
\bar{P}_e = \sum_k p_k^2, \quad p_k = \frac{1}{Nn} \sum_{i} n_{ik},
\label{eq:fleiss}
\end{equation}
where $n_{ik}$ counts the judges who placed battle $i$ in category $k$.

\paragraph{Worked example (illustrative labels).} Consider a redundancy sample of $N = 8$ battles with $n = 3$ judges each and categories $\{\text{A}, \text{B}, \text{tie}\}$, with judgments (A,A,A), (A,A,B), (B,B,B), (A,tie,A), (B,B,tie), (A,A,A), (tie,tie,B), (B,A,B). Three battles are unanimous ($P_i = 1$) and five have a two-to-one split ($P_i = 1/3$), so $\bar{P} = (3 \times 1 + 5 \times 1/3)/8 = 0.583$. The category proportions are $p_{\mathrm{A}} = 11/24 = 0.458$, $p_{\mathrm{B}} = 9/24 = 0.375$, $p_{\mathrm{tie}} = 4/24 = 0.167$, giving $\bar{P}_e = 0.458^2 + 0.375^2 + 0.167^2 = 0.379$ and $\kappa = (0.583 - 0.379)/(1 - 0.379) = 0.330$. The pairwise percent agreement is also $0.583$. A $\kappa$ of $0.33$ is fair agreement by the usual reading and is what one should expect when one judgment in six is a non-preference and five of the eight battles split two to one; the arena publishes both $\kappa$ and the agreement level with ties excluded, because the two answer different questions.

\paragraph{Judge trust.} Each judge carries a graded weight $\omega_v \in [\omega_{\min}, 1]$ computed from two behaviors that cannot be opinions. The first is a position lean: the lower bound of a Wilson interval on the judge's share of votes for the left-hand side, when that interval excludes one half. Sides are shuffled per battle, so position carries no information. The second is a habit of voting faster than the two results could have been read: the lower bound of a Wilson interval on the share of the judge's timed votes that fall under a reading-rate floor. Both factors are $1$ below their evidence floors; the product is clamped to $[\omega_{\min}, 1]$. A judge is marked untrusted, and their votes dropped rather than discounted, only by thresholded evidence: failing gold-standard items with known outcomes, or an accuracy on an elimination task below a chance-corrected floor over enough calls. The number of discounted judges, the number of dropped votes, and the share of the corpus that rests on self-judged battles are published beside the ratings.

\section{Limitations}
\label{sec:limitations}

\paragraph{Human latency is not bounded by the system.} Definition~\ref{def:realtime} excludes $J_b$ on purpose, and the illustrative timeline of Figure~\ref{fig:timeline} makes it the dominant term. An arena can shorten $J_b$ by prioritizing the queue and by recruiting judges; it cannot guarantee a bound. A reader who needs a bound on end-to-end latency should read the reported percentile of $J_b$, not the machine horizon.

\paragraph{Judge population and self-judging.} In a young arena most judgments come from a small number of people, and a large share of battles are judged only by their own submitter. Blinding is enforced by the server for the submitter as for anyone else, so those votes are real blind labels; what they carry is annotator concentration, which the cluster-robust interval widens to reflect and which the published self-judged share makes visible. The intervals are honest about this; the point estimates are still one population's preferences.

\paragraph{Task population.} The tasks are whatever users submit. That is the property that makes the evaluation track use, and it is also a selection effect that no reweighting in this paper corrects. The published distribution over task categories is descriptive; a ranking on it does not transfer to a workload with a different mix.

\paragraph{Ties as half-wins.} The half-win treatment is an approximation. With ties at roughly one vote in seven, as in the worked example (29 of 211), a tie-parameter model \citep{rao1967,davidson1970} would be estimable and would change the intervals; we have not quantified the difference here.

\paragraph{Marginal rank bands.} The rank band uses each rating's marginal variance and ignores the covariance the fit also estimates. The approximation errs toward wider bands, and the direction is stated, but a joint draw from the full covariance would be more accurate.

\section{Conclusion}
\label{sec:conclusion}

We have described an arena for computer-use agents and multi-agent systems in which real tasks are executed by two entrants at once in identical isolated environments, judged blind, and turned into a leaderboard. We have defined what it means for that evaluation to be real-time as five measurable properties: continuous arrival, live concurrent execution, online rating updates, freshness with contamination resistance, and bounded feedback latency to a training environment. Each property has a quantity a deployed system can report at a percentile, and Definition~\ref{def:realtime} names the tier at which each holds. The rating methodology is a penalized Bradley-Terry fit with cluster-robust intervals and bootstrap rank bands, with a streaming Elo update shown to be a first-order online step on the same likelihood. The worked example shows that a few hundred votes separate the ends of a five-agent field and leave neighbors overlapping, which is the honest state of many leaderboards and the reason a rank should be published as a band. The definitions are offered so that the word real-time, when applied to an evaluation, can be checked rather than believed.

\section*{Acknowledgments}
The authors thank the users and judges of CoArena.ai, whose submitted tasks and blind judgments motivated the formal treatment here.

\appendix

\section{Notation}
\label{app:notation}

\begin{table}[H]
\centering
\small
\begin{tabular}{@{}lp{0.72\textwidth}@{}}
\toprule
Symbol & Meaning \\
\midrule
$\Agents$, $m$ & roster of agents and its size \\
$\tau$, $b$, $u$ & a task, a battle (one task, two agents, two runs), a run \\
$s_b, q_b, \delta_b, e_b, h_b, r_b, \ell_b$ & submission, dequeue, provisioning time, execution end, first admissible judgment, rating incorporation, and service times of battle $b$ \\
$L^{\mathrm{q}}_b, L^{\mathrm{p}}_b, L^{\mathrm{e}}_b, L^{\mathrm{r}}_b, L^{\mathrm{s}}_b$ & queue, provisioning, execution, rating-update, and serving latencies~\eqref{eq:latencies} \\
$J_b$ & human judging latency, $h_b - e_b$ \\
$M_b$ & machine latency, the sum of the five machine stages~\eqref{eq:machine} \\
$X_p(S)$ & the $p$-th percentile of latency $X$ over battle set $S$ \\
$N(t)$, $\lambda$, $\lambda(t)$ & arrival counting process, long-run arrival rate, time-varying rate \\
$G_{\max}(T)$ & largest inter-arrival gap on $[0,T]$~\eqref{eq:gap} \\
$C$, $S_b$, $\rho$ & concurrent execution slots, battle service time, offered load~\eqref{eq:utilization} \\
$I(b)$, $\Delta_{\mathrm{start}}$ & resolved input tuple of a battle; bound on lane start skew \\
$K$, $W$ & step budget and wall-clock budget of a run \\
$\varsigma(t)$, $\Delta$, $c$ & rating staleness, the wait of the oldest unincorporated judgment~\eqref{eq:staleness}; refit period; refit compute time \\
$A_b$, $c_b$, $t^{\mathrm{cut}}_a$ & task age at execution, contamination indicator~\eqref{eq:contam}, training cutoff of agent $a$ \\
$F(\alpha;t)$, $N_w$ & share of rating weight from tasks newer than $\alpha$~\eqref{eq:freshness}; fitting window in battles \\
$D(\theta)$, $\mathrm{sim}$ & duplicate rate at similarity threshold $\theta$~\eqref{eq:duplicate}; lexical similarity \\
$\mathcal{F}$, $\varepsilon_u$, $\nu_u$, $T^{\mathrm{env}}_u$ & failed runs, derived environment, its publication time, feedback latency~\eqref{eq:tenv} \\
$H$, $H_{\mathrm{exec}}, H_{\mathrm{rate}}, H_{\mathrm{age}}, H_{\mathrm{env}}$ & real-time horizons \\
\midrule
$\theta_a$, $\theta$, $\hat{\theta}$ & latent strength of agent $a$, the vector of strengths, its estimate \\
$R_a$, $R_0$ & rating on the Elo scale~\eqref{eq:eloscale} and its origin, $1000$ \\
$p_{ij}$, $\sig(\cdot)$ & preference probability~\eqref{eq:bt}; logistic function \\
$y_b$, $x_b$, $\eta_b$, $\mu_b$ & outcome, design vector, linear predictor, fitted probability of battle $b$ \\
$w_b$, $\gamma_b$, $\omega_v$, $\omega_{\min}$ & evidence weight~\eqref{eq:weight}, calibration discount, judge trust weight, its floor \\
$\mathcal{L}(\theta)$, $r$ & log-likelihood~\eqref{eq:loglik}; ridge penalty \\
$g(\theta)$, $H(\theta)$ & gradient and negative Hessian of the penalized objective~\eqref{eq:gradhess} \\
$P$ & centering projection $I - \mathbf{1}\mathbf{1}^{\top}/m$ \\
$G$, $u_g$ & number of judge clusters; summed score of cluster $g$~\eqref{eq:sandwich} \\
$K_a$, $K_{\max}, K_{\min}, n_0$ & streaming factor of agent $a$ and its schedule~\eqref{eq:kfactor}; provisional floor \\
$n_a$, $\kappa$ & rated battles of agent $a$; matchmaking pseudo-count \\
$\bar{P}$, $\bar{P}_e$, $\kappa$ (agreement) & observed and chance agreement, Fleiss' kappa~\eqref{eq:fleiss} \\
\midrule
$\epsilon$, $T_{\max}$ & Newton tolerance and iteration cap (Algorithm~\ref{alg:refit}) \\
$z$, $\psi$ & Newton step and per-battle working weight $w_b \mu_b (1-\mu_b)$ (Algorithm~\ref{alg:refit}) \\
$\Sigma$, $\mathrm{SE}(\cdot)$ & estimated covariance of $\hat{\theta}$ and a standard error (Algorithm~\ref{alg:refit}) \\
$\mathcal{P}$, $B$, $\zeta_0$, $\xi_a$ & publishable agents, bootstrap draw count, base seed, and agent $a$'s seeded stream (Algorithm~\ref{alg:bands}) \\
$\tilde{R}^{(d)}_a$, $\mathrm{pos}_d(a)$ & agent $a$'s $d$-th drawn rating and its position in that draw (Algorithm~\ref{alg:bands}) \\
$t_0$ & failure time from which the feedback latency is measured, $e_u$ (Algorithm~\ref{alg:pipeline}) \\
\bottomrule
\end{tabular}
\caption{Notation used in the paper, in order of first appearance. The symbol $\kappa$ is used for the matchmaking pseudo-count in Section~\ref{sec:system} and for Fleiss' agreement coefficient in Section~\ref{sec:votequality}; the context makes the meaning unambiguous in each case.}
\label{tab:notation}
\end{table}

\section{Algorithms}
\label{app:algorithms}

\begin{algorithm}[H]
\caption{Streaming update on one resolved judgment (Section~\ref{sec:streaming})}
\label{alg:stream}
\begin{algorithmic}[1]
\Require ratings $R$, rated-battle counts $n$, battle $b$ with agents $(i, j)$, outcome $y_b \in \{1, \tfrac12, 0\}$, weight $w_b$
\State $\mu \gets 1 / \left(1 + 10^{(R_j - R_i)/400}\right)$ \Comment{expected score of $i$, equation~\eqref{eq:elo}}
\State $K_i \gets K_{\min} + (K_{\max} - K_{\min})\, n_0 / (n_0 + n_i)$ \Comment{equation~\eqref{eq:kfactor}}
\State $K_j \gets K_{\min} + (K_{\max} - K_{\min})\, n_0 / (n_0 + n_j)$
\State $R_i \gets R_i + K_i\, w_b\, (y_b - \mu)$
\State $R_j \gets R_j - K_j\, w_b\, (y_b - \mu)$
\State $n_i \gets n_i + 1$; \; $n_j \gets n_j + 1$
\State \Return $R$, $n$
\end{algorithmic}
\end{algorithm}

\begin{algorithm}[H]
\caption{Periodic refit: penalized maximum likelihood with projected covariance (Section~\ref{sec:estimator})}
\label{alg:refit}
\begin{algorithmic}[1]
\Require admissible battles $\{(i_b, j_b, y_b, w_b)\}_{b=1}^{n}$, penalty $r$, tolerance $\epsilon$, maximum iterations $T_{\max}$
\State $\theta \gets 0 \in \Real^m$
\For{$t = 1, \ldots, T_{\max}$}
  \State $g \gets -r\theta$; \; $H \gets rI$
  \For{each battle $b$}
    \State $\mu_b \gets \sig(\theta_{i_b} - \theta_{j_b})$; \; $\psi \gets w_b\, \mu_b (1 - \mu_b)$
    \State $g_{i_b} \gets g_{i_b} + w_b (y_b - \mu_b)$; \; $g_{j_b} \gets g_{j_b} - w_b (y_b - \mu_b)$
    \State $H_{i_b i_b} \gets H_{i_b i_b} + \psi$; \; $H_{j_b j_b} \gets H_{j_b j_b} + \psi$; \; $H_{i_b j_b} \gets H_{i_b j_b} - \psi$; \; $H_{j_b i_b} \gets H_{j_b i_b} - \psi$
  \EndFor
  \State $z \gets H^{-1} g$; \; $\theta \gets \theta + z$
  \If{$\|z\|_\infty < \epsilon$} \textbf{break} \EndIf
\EndFor
\State $\bar{\theta} \gets$ mean of $\theta_a$ over agents with $n_a \ge n_0$ (all agents if none has reached $n_0$)
\State $\theta \gets \theta - \bar{\theta}\,\mathbf{1}$ \Comment{center on the settled field}
\State $\Sigma \gets P H^{-1} P$ with $P = I - \mathbf{1}\mathbf{1}^{\top}/m$ \Comment{equation~\eqref{eq:modelci}}
\State optionally replace $\Sigma$ by the cluster-robust estimate~\eqref{eq:sandwich} when the number of judge clusters exceeds $m$
\State $R_a \gets R_0 + (400/\ln 10)\,\theta_a$; \; $\mathrm{SE}(R_a) \gets (400/\ln 10)\sqrt{\Sigma_{aa}}$ for each $a$
\State \Return $R$, $\mathrm{SE}$, the set of agents with $n_a \ge n_0$ (publishable)
\end{algorithmic}
\end{algorithm}

\begin{algorithm}[H]
\caption{Rank bands by seeded parametric bootstrap (Section~\ref{sec:intervals})}
\label{alg:bands}
\begin{algorithmic}[1]
\Require publishable agents $\mathcal{P} \subseteq \Agents$ with ratings $R_a$ and standard errors $\mathrm{SE}(R_a)$, draw count $B$, base seed $\zeta_0$
\For{each $a \in \mathcal{P}$}
  \State initialize a pseudo-random stream $\xi_a$ from $\mathrm{hash}(a) \oplus \zeta_0$ \Comment{per-agent, order-independent}
  \State draw $\tilde{R}_a^{(1)}, \ldots, \tilde{R}_a^{(B)} \sim \mathcal{N}(R_a, \mathrm{SE}(R_a)^2)$ from $\xi_a$
\EndFor
\State $\mathrm{count}_a[k] \gets 0$ for all $a$ and positions $k$
\For{$d = 1, \ldots, B$}
  \State sort $\mathcal{P}$ by $\tilde{R}_a^{(d)}$ descending; let $\mathrm{pos}_d(a)$ be $a$'s position
  \State $\mathrm{count}_a[\mathrm{pos}_d(a)] \gets \mathrm{count}_a[\mathrm{pos}_d(a)] + 1$ for each $a$
\EndFor
\For{each $a \in \mathcal{P}$}
  \State $\mathrm{low}_a \gets$ smallest $k$ with cumulative count reaching $0.025\,B$; \; $\mathrm{high}_a \gets$ smallest $k$ with cumulative count reaching $0.975\,B$
  \State $\mathrm{point}_a \gets 1 + \#\{a' \in \mathcal{P} : R_{a'} > R_a\}$ \Comment{ties share a position}
  \State $\mathrm{low}_a \gets \min(\mathrm{low}_a, \mathrm{point}_a)$; \; $\mathrm{high}_a \gets \max(\mathrm{high}_a, \mathrm{point}_a)$
  \State $\Prob\{a \text{ first}\} \gets \mathrm{count}_a[1] / B$
\EndFor
\State \Return $(\mathrm{point}_a, \mathrm{low}_a, \mathrm{high}_a, \Prob\{a \text{ first}\})$ for each $a \in \mathcal{P}$
\end{algorithmic}
\end{algorithm}

\begin{algorithm}[H]
\caption{Failure-to-environment pipeline (Section~\ref{sec:feedback})}
\label{alg:pipeline}
\begin{algorithmic}[1]
\Require failed run $u$ with trajectory, frames, initial snapshot, instruction, outcome; horizon $H_{\mathrm{env}}$
\State $t_0 \gets e_u$
\State \textbf{capture}: persist trajectory, frames, and the pre-first-action snapshot of the sandbox
\State \textbf{redact}: remove personal data from the instruction, frames, and typed text; abort if redaction is incomplete
\State \textbf{synthesize}: $\varepsilon_u \gets$ (snapshot, redacted instruction, grader) where the grader is derived from the failure mode
\State \textbf{validate}: replay $u$'s trajectory in $\varepsilon_u$; require the failure to reproduce
\State \textbf{validate}: run a reference trajectory in $\varepsilon_u$; require the grader to pass it
\If{either validation fails} \State \Return rejected with reason \EndIf
\State \textbf{publish} $\varepsilon_u$ to the registry at time $\nu_u$
\State $T^{\mathrm{env}}_u \gets \nu_u - t_0$; record it; flag if $T^{\mathrm{env}}_u > H_{\mathrm{env}}$
\State \Return $\varepsilon_u$, $T^{\mathrm{env}}_u$
\end{algorithmic}
\end{algorithm}

\section{Interface Walkthrough and a Worked Session}
\label{app:walkthrough}

This appendix shows the four interfaces a participant meets in CoArena and carries one battle through all of them with the numbers of Table~\ref{tab:worked}. The interface figures are schematic renderings drawn to show the fields and their order; they are not captured screenshots, and every number on them is either taken from Table~\ref{tab:worked}, derived in Section~\ref{app:session}, or labeled illustrative. The tasks in Table~\ref{tab:tasks} are written for this paper to show the range a deployed arena receives; the tasks of a deployed arena are whatever its users submit.

\subsection{Example tasks}

\begin{table}[H]
\centering
\small
\begin{tabular}{@{}p{0.31\textwidth}p{0.13\textwidth}p{0.24\textwidth}p{0.24\textwidth}@{}}
\toprule
Task as submitted (illustrative) & Category & What the judge compares & Failure mode and derived grader \\
\midrule
Find a nonstop economy fare from San Francisco to New York on the 14th for one passenger, stop before payment, and report the fare. & web research, form filling & Whether each run honored every constraint (nonstop, date, one passenger), stopped before payment, and reported a fare the page actually showed. & A run selects a one-stop fare or proceeds to payment. Grader: delivered fare must match a nonstop itinerary in the final state, and no payment step may appear in the trajectory. \\
\addlinespace
From the attached receipts (CSV), fill the expense form in the browser and download the summary as a PDF. & form filling, attachments & Whether the downloaded PDF exists, whether its totals match the CSV, and how many receipts were entered correctly. & A run exhausts its step budget with rows missing. Grader: parse the delivered PDF and compare totals and row count with the attachment. \\
\addlinespace
Open the spreadsheet on the desktop, add a column with each row's share of the total, and save it under a new name. & desktop application & Whether the new file exists, whether the column is arithmetically right, and whether the original is unchanged. & A run overwrites the original. Grader: original hash unchanged, new file present, column values within rounding of the expected shares. \\
\addlinespace
Rename every file in the Downloads folder with a date prefix taken from its modification time and archive them into one zip. & desktop, file operations & Whether every file was renamed by the stated rule and whether the archive contains exactly those files. & A run renames some files and stops. Grader: enumerate the folder and the archive in the final state; every name must match the rule. \\
\addlinespace
Create a weekly recurring meeting in the calendar application with two named attendees and send the invitation. & desktop application & Whether the event recurs weekly, whether both attendees are attached, and whether an invitation was sent. & A run creates a one-time event. Grader: read the event record in the final state and check recurrence and attendee fields. \\
\addlinespace
Compare the return policies on two given retailer pages and write a one-paragraph summary of the differences that matter to a buyer. & web research, writing & Which summary is accurate to the pages and more useful to the person asking; this is a preference comparison rather than a checklist. & A run summarizes only one page. Grader: the summary must cite facts present on both pages; the judgment remains a preference. \\
\bottomrule
\end{tabular}
\caption{Example tasks written for this paper to show the range a computer-use arena receives, with the category the intake assigns, what a blind judge compares between the two runs, and how a failed run converts into an environment with a grader derived from the failure (Section~\ref{sec:feedback}). None of these is a measured task from a deployed arena.}
\label{tab:tasks}
\end{table}

\subsection{The leaderboard}

Figure~\ref{fig:ui-board} renders the leaderboard as a participant sees it, populated with the five agents of Table~\ref{tab:worked}. The Rating column carries the refit's point estimate, with its 95\% interval in the next column; the rank column carries the band of Algorithm~\ref{alg:bands}, printed as a range wherever the data cannot pin a position, and the status column states which of the four evidence levels the row has reached. A sixth, provisional row is included to show how an agent below the floor of Section~\ref{sec:entry} is listed: it sinks below every rated row, shows the reason its number is withheld, and takes no rank.

\begin{figure}[H]
\centering
\resizebox{0.96\textwidth}{!}{%
\begin{tikzpicture}[font=\small, x=1cm, y=1cm]
\def\W{15.2}
\draw[rounded corners=3pt, thick] (0,0) rectangle (\W,-8.75);
\fill[gray!12, rounded corners=3pt] (0,0) rectangle (\W,-0.9);
\node[anchor=west, font=\small\bfseries] at (0.3,-0.45) {Leaderboard};
\node[anchor=west, font=\scriptsize] at (3.0,-0.45) {Rating \; $|$ \; Success \; $|$ \; Speed \; $|$ \; Effort \; $|$ \; Cost};
\node[anchor=north west, font=\scriptsize, text width=14.6cm, align=left] at (0.3,-1.05) {Blind human evaluation, and nothing else: judges pick the better of two runs without knowing which model made them. Refit every 30 s; the streaming value is shown in each model card.};
\fill[gray!6] (0,-1.85) rectangle (\W,-2.35);
\foreach \x/\t in {0.3/Rank, 1.6/Agent, 6.0/Rating, 8.3/95\% interval, 10.6/Status, 12.6/Judgments} {\node[anchor=west, font=\scriptsize\bfseries] at (\x,-2.1) {\t};}
\foreach \y/\r/\a/\v/\ci/\st/\n in {
  -2.9/1 to 2/Agent A/1140/1073 to 1206/ranked/86,
  -3.6/1 to 3/Agent B/1050/988 to 1111/ranked/87,
  -4.3/2 to 4/Agent C/992/931 to 1053/ranked/87,
  -5.0/3 to 5/Agent D/941/877 to 1005/ranked/82,
  -5.7/4 to 5/Agent E/877/809 to 945/ranked/80} {
  \node[anchor=west, font=\scriptsize] at (0.3,\y) {\r};
  \node[anchor=west, font=\scriptsize] at (1.6,\y) {\a};
  \node[anchor=west, font=\scriptsize] at (6.0,\y) {\v};
  \node[anchor=west, font=\scriptsize] at (8.3,\y) {\ci};
  \node[anchor=west, font=\scriptsize] at (10.6,\y) {\st};
  \node[anchor=west, font=\scriptsize] at (12.6,\y) {\n};
  \draw[gray!30] (0.2,\y-0.32) to (\W-0.2,\y-0.32);
}
\node[anchor=west, font=\scriptsize] at (1.6,-6.4) {Agent F (illustrative)};
\node[anchor=west, font=\scriptsize, gray!60!black] at (6.0,-6.4) {under 30 comparisons};
\node[anchor=west, font=\scriptsize, gray!60!black] at (10.6,-6.4) {provisional};
\node[anchor=west, font=\scriptsize, gray!60!black] at (12.6,-6.4) {7};
\draw[gray!30] (0.2,-6.72) to (\W-0.2,-6.72);
\node[anchor=north west, font=\scriptsize, text width=14.6cm, align=left] at (0.3,-6.9) {Rank is a 95\% band from the fit's own uncertainty: a range means the data cannot separate those positions. Ratings are a Bradley-Terry fit over every admissible blind judgment, with standard errors clustered on the judge. An agent with fewer than 30 rated comparisons appears in the list without a rating. Judgment counts are shown for the worked example only; a public board may withhold volume.};
\end{tikzpicture}%
}
\caption{The leaderboard, rendered schematically with the five agents of Table~\ref{tab:worked} and one illustrative provisional row. The figure shows that the rank column is a band rather than an ordinal (A holds 1 to 2, B holds 1 to 3, and neither is printed as first), that the rating and its interval come from the same refit, and that a provisional agent is listed with the reason its number is withheld rather than ranked on a ticker.}
\label{fig:ui-board}
\end{figure}

\subsection{The judging view}

Figure~\ref{fig:ui-judge} renders the view a judge sees for the first task of Table~\ref{tab:tasks}. The two runs are labeled by side only. The judge can replay each trajectory step by step, read each final message, open each delivered file, and then choose one of the five responses of Section~\ref{sec:votequality}. Identities are withheld by the server until the vote is recorded; the interface has nothing to reveal early.

\begin{figure}[H]
\centering
\resizebox{0.96\textwidth}{!}{%
\begin{tikzpicture}[font=\small, x=1cm, y=1cm]
\def\W{15.2}
\draw[rounded corners=3pt, thick] (0,0) rectangle (\W,-9.45);
\fill[gray!12, rounded corners=3pt] (0,0) rectangle (\W,-0.9);
\node[anchor=west, font=\small\bfseries] at (0.3,-0.45) {Judge this battle};
\node[anchor=east, font=\scriptsize] at (\W-0.3,-0.45) {identities withheld until you vote \; $\cdot$ \; 1 of 3 judges on this battle};
\node[anchor=north west, font=\scriptsize, text width=14.6cm, align=left] at (0.3,-1.05) {\textbf{Task.} Find a nonstop economy fare from San Francisco to New York on the 14th for one passenger, stop before payment, and report the fare.};
\foreach \x/\side in {0.3/Left run, 7.85/Right run} {
  \draw[rounded corners=2pt] (\x,-2.0) rectangle (\x+7.05,-7.2);
  \node[anchor=west, font=\scriptsize\bfseries] at (\x+0.15,-2.25) {\side};
  \node[anchor=east, font=\scriptsize] at (\x+6.9,-2.25) {replay \; $\triangleright$};
  \foreach \k in {0,1,2,3,4} {
    \fill[gray!18] (\x+0.2+\k*1.36,-2.55) rectangle (\x+0.2+\k*1.36+1.2,-3.45);
    \node[font=\tiny] at (\x+0.8+\k*1.36,-3.0) {step \k};
  }
  \node[anchor=north west, font=\tiny, gray!60!black, text width=6.7cm, align=left] at (\x+0.15,-3.55) {screenshot on every step; actions shown beneath each frame};
}
\node[anchor=north west, font=\scriptsize, align=left, text width=6.7cm] at (0.45,-4.2) {\textbf{Final message.} Nonstop economy fare found for the 14th, one passenger: \$248 on the 7:05 departure. Stopped at the payment step as instructed.\\[2pt]\textbf{Delivered.} fare-summary.txt};
\node[anchor=north west, font=\scriptsize, align=left, text width=6.7cm] at (8.0,-4.2) {\textbf{Final message.} Lowest fare for the 14th is \$219 (one stop). Proceeded to the payment page and stopped before entering card details.\\[2pt]\textbf{Delivered.} none};
\node[anchor=west, font=\scriptsize\bfseries] at (0.3,-7.6) {Which run did the task better?};
\foreach \x/\lab in {0.3/Left is better, 3.2/Tie, 6.1/Right is better, 9.0/Both unacceptable, 11.9/Skip this battle} {
  \draw[rounded corners=2pt] (\x,-7.9) rectangle ++(2.7,-0.6);
  \node[font=\scriptsize] at (\x+1.35,-8.2) {\lab};
}
\node[anchor=north west, font=\tiny, gray!60!black, text width=14.6cm, align=left] at (0.3,-8.65) {Both runs received the same instruction, attachments, tools and budgets, and ran at the same time in separate isolated desktops.};
\end{tikzpicture}%
}
\caption{The judging view, rendered schematically for the first task of Table~\ref{tab:tasks} with illustrative run contents. What the figure shows is that a judge compares two complete runs of the same task, side by side and blind, with the trajectory, the final message and the delivered files all available; that the five responses map onto the outcome codes of Section~\ref{sec:observations} (a side, a tie, both unacceptable, or an abstention that returns the battle to the queue); and that in this instance the right-hand run violated the nonstop constraint and delivered nothing, which is the kind of difference the preference label is meant to capture.}
\label{fig:ui-judge}
\end{figure}

\subsection{The model card}

Figure~\ref{fig:ui-card} renders the card that opens from an agent's name on the board, for Agent A of Table~\ref{tab:worked}. It carries the same rating and interval as the board row, the rank band with the probability of first place from Algorithm~\ref{alg:bands}, the estimator that produced the number, and the run-level summaries. The streaming value is shown under its own name; it is what moved when the judge of Figure~\ref{fig:ui-judge} voted, and the card says so.

\begin{figure}[H]
\centering
\resizebox{0.8\textwidth}{!}{%
\begin{tikzpicture}[font=\small, x=1cm, y=1cm]
\def\W{12.6}
\draw[rounded corners=3pt, thick] (0,0) rectangle (\W,-7.6);
\fill[gray!12, rounded corners=3pt] (0,0) rectangle (\W,-0.8);
\node[anchor=west, font=\small\bfseries] at (0.3,-0.4) {Agent A};
\node[anchor=east, font=\scriptsize] at (\W-0.3,-0.4) {model card};
\node[anchor=west, font=\scriptsize, gray!60!black] at (0.3,-1.2) {Rating};
\node[anchor=west, font=\Large\bfseries] at (0.3,-1.85) {1140};
\node[anchor=west, font=\small] at (1.75,-1.85) {$\pm 67$};
\node[anchor=west, font=\scriptsize] at (2.7,-1.85) {arena};
\node[anchor=north west, font=\scriptsize, text width=7.7cm, align=left] at (0.3,-2.2) {Bradley-Terry over 86 admissible blind judgments; standard errors clustered on the judge.};
\node[anchor=north west, font=\scriptsize, text width=7.7cm, align=left] at (0.3,-2.95) {Streaming value: 1148 (moves when a judge votes; not the published rating).};
\draw[rounded corners=2pt] (8.5,-1.1) rectangle (\W-0.3,-3.5);
\node[anchor=west, font=\scriptsize\bfseries] at (8.65,-1.45) {rank band 1 to 2};
\node[anchor=west, font=\scriptsize] at (8.65,-1.95) {1st of 5 ranked models};
\node[anchor=west, font=\scriptsize] at (8.65,-2.45) {97.3\% chance of first};
\node[anchor=west, font=\scriptsize] at (8.65,-2.95) {status: ranked};
\draw[gray!30] (0.2,-3.85) to (\W-0.2,-3.85);
\foreach \x/\lab/\v in {0.3/Score/0.721, 3.4/{Wins, losses, ties}/{57, 19, 10}, 7.2/Completion/[measured], 10.0/Median run/[measured]} {
  \node[anchor=west, font=\scriptsize, gray!60!black] at (\x,-4.2) {\lab};
  \node[anchor=west, font=\small] at (\x,-4.65) {\v};
}
\draw[gray!30] (0.2,-5.05) to (\W-0.2,-5.05);
\node[anchor=north west, font=\scriptsize, text width=12.0cm, align=left] at (0.3,-5.25) {The rating is fitted from blind preference votes alone and refit from scratch on every refresh; its 95\% interval accounts for votes grouped by judge. The rank band is the range of positions this model held across 2,000 draws of every rating from its own interval.};
\node[anchor=north west, font=\scriptsize, text width=12.0cm, align=left, gray!60!black] at (0.3,-6.55) {Score, wins, losses and ties are the worked example's values. Completion and median run are run-level summaries a deployed card measures; they are placeholders here.};
\end{tikzpicture}%
}
\caption{The model card for Agent A, rendered schematically from Table~\ref{tab:worked}. The card, as the figure shows, prints the same rating and interval as the board row, states the estimator and its clustering, gives the rank as a band with the probability of first place rather than as an ordinal, and shows the streaming value under its own name so that a judge who has just voted can see what moved without mistaking it for the published number. The streaming value shown is the result of the worked session in Section~\ref{app:session}.}
\label{fig:ui-card}
\end{figure}

\subsection{An environment record}

Figure~\ref{fig:ui-env} renders the registry entry produced when the right-hand run of Figure~\ref{fig:ui-judge} is converted by the pipeline of Section~\ref{sec:feedback}. The run failed the nonstop constraint, so the derived grader checks the itinerary type in the final state and the absence of a payment step in the trajectory. The stage times are the illustrative values of Section~\ref{sec:feedback}.

\begin{figure}[H]
\centering
\resizebox{0.8\textwidth}{!}{%
\begin{tikzpicture}[font=\small, x=1cm, y=1cm]
\def\W{12.6}
\draw[rounded corners=3pt, thick] (0,0) rectangle (\W,-9.3);
\fill[gray!12, rounded corners=3pt] (0,0) rectangle (\W,-0.8);
\node[anchor=west, font=\small\bfseries] at (0.3,-0.4) {Environment record};
\node[anchor=east, font=\scriptsize] at (\W-0.3,-0.4) {validated \; $\cdot$ \; published};
\foreach \y/\k/\v in {
  -1.1/Source/{right-hand run of the battle in Figure~\ref{fig:ui-judge}; the agent's identity is withheld here as on the board},
  -1.85/Failure mode/{constraint violated: the itinerary had one stop; the task required nonstop},
  -2.6/Initial state/{desktop snapshot taken before the first action; content hash [stored]},
  -3.35/Instruction/{the task text after redaction; no personal data remained to remove},
  -4.1/Grader/{final state must show a nonstop itinerary; the trajectory must contain no payment step},
  -4.85/Reference/{the left-hand run's trajectory passes the grader},
  -5.6/Replay check/{the failing trajectory reproduces the failure in this environment},
  -6.35/Stage times/{capture 20 s $\cdot$ redaction 40 s $\cdot$ synthesis 300 s $\cdot$ validation 960 s $\cdot$ publish 10 s (illustrative)},
  -7.1/Latency/{$T^{\mathrm{env}} = 1{,}330$ s $\approx 22$ min against a horizon $H_{\mathrm{env}} = 1$ h: within horizon}} {
  \node[anchor=north west, font=\scriptsize\bfseries] at (0.3,\y) {\k};
  \node[anchor=north west, font=\scriptsize, text width=9.3cm, align=left] at (2.9,\y) {\v};
}
\draw[gray!30] (0.2,-7.95) to (\W-0.2,-7.95);
\node[anchor=north west, font=\scriptsize, text width=12.0cm, align=left] at (0.3,-8.15) {A record is published only after both validations pass. Developers pull the initial state, the instruction and the grader; the trajectory that produced the failure is included as a reference of what to avoid, with the agent's identity withheld.};
\end{tikzpicture}%
}
\caption{An environment record in the registry, rendered schematically for the failed run of Figure~\ref{fig:ui-judge}. The record shows that an environment is a triple of initial state, instruction and grader, that the grader is derived from the specific way the run failed, that publication is conditional on the failing trajectory reproducing its failure and a reference passing, and that the record carries its own feedback latency against the stated horizon.}
\label{fig:ui-env}
\end{figure}

\subsection{One battle, end to end}
\label{app:session}

The following walks one battle through the system with the illustrative event times of Figure~\ref{fig:timeline} and the ratings of Table~\ref{tab:worked}. The two agents drawn are A and B.

\begin{enumerate}
\item \textbf{Submission} ($s_b = 0$). A user submits the first task of Table~\ref{tab:tasks}. Intake authenticates the user, confirms consent, classifies the text as a task, finds no near-duplicate among the user's recent tasks, and admits it. Elapsed: under one second.
\item \textbf{Matchmaking and dequeue} ($q_b = 5$ s). The sampler draws Agents A and B. With 86 and 87 rated battles they are the two agents with the most accumulated evidence, and the pairing is informative: $\sig(\hat{\theta}_A - \hat{\theta}_B) = \sig(0.804 - 0.288) = 0.626$ is well inside the range where a comparison carries information. The battle waits 5 s for an execution slot.
\item \textbf{Provisioning and lane start} ($q_b + \delta_b = 45$ s). Two isolated desktops are provisioned. Both lanes receive the same instruction, the same tool schemas, the same step budget $K$ and wall-clock budget $W$, and start within the skew bound.
\item \textbf{Execution} ($e_b = 6$ min 45 s). Agent A finds a nonstop fare and stops before payment; Agent B selects a one-stop fare and reaches the payment page. Both runs end within the budgets. Every step's screenshot and action are recorded for both.
\item \textbf{Judging} ($h_b = 3$ h 6 min 45 s). The battle was drawn for redundancy at creation, so it collects three judges. The first arrives after about three hours (illustrative) and votes for the left-hand run, which is Agent A. The two later judges also vote A; the consensus is A with three agreeing judges, and the battle contributes $P_i = 1$ to the agreement statistic of Section~\ref{sec:votequality}.
\item \textbf{Streaming update} (within milliseconds of the first vote). Before the vote, $R_A = 1139.7$ and $R_B = 1050.0$. By equation~\eqref{eq:elo}, the expected score of A is $\mu = 1 / (1 + 10^{(1050.0 - 1139.7)/400}) = 1/(1 + 10^{-0.2243}) = 0.6263$. The factors from equation~\eqref{eq:kfactor} are $K_A = 12 + 36 \cdot 30/(30 + 86) = 21.31$ and $K_B = 12 + 36 \cdot 30/(30 + 87) = 21.23$. With $y_b = 1$ and $w_b = 1$, A gains $K_A (1 - \mu) = 7.96$ points and B loses $K_B (1 - \mu) = 7.93$ points: the streaming values become $R_A = 1147.7$ and $R_B = 1042.1$. The judge sees these move at once, labeled as the streaming value.
\item \textbf{Refit} (within 31 s of the vote). The next periodic refit runs Algorithm~\ref{alg:refit} on the 212 admissible battles. Adding one A-over-B outcome to the 211 of Table~\ref{tab:worked} moves the centered estimates to $R_A = 1142.6$ (a change of $+2.9$) and $R_B = 1047.5$ (a change of $-2.5$); the interval half-widths change by less than one Elo point. The published rating and its band are recomputed from this fit, and the board that clients poll reflects the vote with staleness under 31 s.
\item \textbf{Feedback} ($\nu_u \approx e_b + 22$ min, in parallel with judging). Agent B's run ended with a constraint violation, so the pipeline of Section~\ref{sec:feedback} captured, redacted, synthesized, validated and published the environment record of Figure~\ref{fig:ui-env} about 22 minutes after the run ended, well before the first judge arrived. This was possible because the failure was a constraint violation a checker can detect without a judgment, so the grader was derived from the constraint and the left-hand run served as the mechanical reference; a wrong-result failure whose grader needs the judged-correct run waits for the judgment.
\end{enumerate}

Two things in this walk-through are the point of the paper. The streaming update and the refit agree in direction and differ in magnitude, because one is a single first-order step and the other re-solves the whole likelihood; the board publishes the second and shows the first. And the machine stages, from submission to a published refit, total about seven minutes plus 31 s, while the human stage took three hours: the system can bound what it controls, and it reports what it does not.


\begin{thebibliography}{99}

\bibitem{boubdir2023elo}
M. Boubdir, E. Kim, B. Ermis, S. Hooker, and M. Fadaee.
Elo uncovered: Robustness and best practices in language model evaluation.
In \emph{Proceedings of the Third Workshop on Natural Language Generation, Evaluation, and Metrics (GEM)}, pages 339 to 352, 2023.

\bibitem{bradley1952}
R. A. Bradley and M. E. Terry.
Rank analysis of incomplete block designs: I. The method of paired comparisons.
\emph{Biometrika}, 39(3/4):324 to 345, 1952.

\bibitem{chiang2024chatbot}
W.-L. Chiang, L. Zheng, Y. Sheng, A. N. Angelopoulos, T. Li, D. Li, B. Zhu, H. Zhang, M. Jordan, J. E. Gonzalez, and I. Stoica.
Chatbot Arena: An open platform for evaluating LLMs by human preference.
In \emph{Proceedings of the 41st International Conference on Machine Learning (ICML)}, volume 235 of \emph{Proceedings of Machine Learning Research}, pages 8359 to 8388, 2024.

\bibitem{cohen1960}
J. Cohen.
A coefficient of agreement for nominal scales.
\emph{Educational and Psychological Measurement}, 20(1):37 to 46, 1960.

\bibitem{davidson1970}
R. R. Davidson.
On extending the Bradley-Terry model to accommodate ties in paired comparison experiments.
\emph{Journal of the American Statistical Association}, 65(329):317 to 328, 1970.

\bibitem{deng2023mind2web}
X. Deng, Y. Gu, B. Zheng, S. Chen, S. Stevens, B. Wang, H. Sun, and Y. Su.
Mind2Web: Towards a generalist agent for the web.
In \emph{Advances in Neural Information Processing Systems 36 (NeurIPS), Datasets and Benchmarks Track}, 2023.

\bibitem{efron1979}
B. Efron.
Bootstrap methods: Another look at the jackknife.
\emph{The Annals of Statistics}, 7(1):1 to 26, 1979.

\bibitem{elo1978}
A. E. Elo.
\emph{The Rating of Chessplayers, Past and Present}.
Arco Publishing, New York, 1978.

\bibitem{fleiss1971}
J. L. Fleiss.
Measuring nominal scale agreement among many raters.
\emph{Psychological Bulletin}, 76(5):378 to 382, 1971.

\bibitem{ford1957}
L. R. Ford, Jr.
Solution of a ranking problem from binary comparisons.
\emph{The American Mathematical Monthly}, 64(8, Part 2):28 to 33, 1957.

\bibitem{glickman1999}
M. E. Glickman.
Parameter estimation in large dynamic paired comparison experiments.
\emph{Journal of the Royal Statistical Society: Series C (Applied Statistics)}, 48(3):377 to 394, 1999.

\bibitem{herbrich2006trueskill}
R. Herbrich, T. Minka, and T. Graepel.
TrueSkill: A Bayesian skill rating system.
In \emph{Advances in Neural Information Processing Systems 19 (NIPS)}, 2006.

\bibitem{hunter2004}
D. R. Hunter.
MM algorithms for generalized Bradley-Terry models.
\emph{The Annals of Statistics}, 32(1):384 to 406, 2004.

\bibitem{jacovi2023stop}
A. Jacovi, A. Caciularu, O. Goldman, and Y. Goldberg.
Stop uploading test data in plain text: Practical strategies for mitigating data contamination by evaluation benchmarks.
In \emph{Proceedings of the 2023 Conference on Empirical Methods in Natural Language Processing (EMNLP)}, 2023.

\bibitem{jain2024livecodebench}
N. Jain, K. Han, A. Gu, W.-D. Li, F. Yan, T. Zhang, S. Wang, A. Solar-Lezama, K. Sen, and I. Stoica.
LiveCodeBench: Holistic and contamination free evaluation of large language models for code.
In \emph{Proceedings of the 13th International Conference on Learning Representations (ICLR)}, 2025.

\bibitem{kiela2021dynabench}
D. Kiela, M. Bartolo, Y. Nie, D. Kaushik, A. Geiger, Z. Wu, B. Vidgen, G. Prasad, A. Singh, P. Ringshia, Z. Ma, T. Thrush, S. Riedel, Z. Waseem, P. Stenetorp, R. Jia, M. Bansal, C. Potts, and A. Williams.
Dynabench: Rethinking benchmarking in NLP.
In \emph{Proceedings of the 2021 Conference of the North American Chapter of the Association for Computational Linguistics: Human Language Technologies (NAACL-HLT)}, pages 4110 to 4124, 2021.

\bibitem{kingman1993poisson}
J. F. C. Kingman.
\emph{Poisson Processes}.
Oxford University Press, Oxford, 1993.

\bibitem{koh2024visualwebarena}
J. Y. Koh, R. Lo, L. Jang, V. Duvvur, M. C. Lim, P.-Y. Huang, G. Neubig, S. Zhou, R. Salakhutdinov, and D. Fried.
VisualWebArena: Evaluating multimodal agents on realistic visual web tasks.
In \emph{Proceedings of the 62nd Annual Meeting of the Association for Computational Linguistics (ACL)}, 2024.

\bibitem{krippendorff2018}
K. Krippendorff.
\emph{Content Analysis: An Introduction to Its Methodology}.
SAGE Publications, Thousand Oaks, fourth edition, 2018.

\bibitem{li2024arenahard}
T. Li, W.-L. Chiang, E. Frick, L. Dunlap, T. Wu, B. Zhu, J. E. Gonzalez, and I. Stoica.
From crowdsourced data to high-quality benchmarks: Arena-Hard and BenchBuilder pipeline.
In \emph{Proceedings of the 42nd International Conference on Machine Learning (ICML)}, volume 267 of \emph{Proceedings of Machine Learning Research}, pages 34209 to 34231, 2025.

\bibitem{liang1986}
K.-Y. Liang and S. L. Zeger.
Longitudinal data analysis using generalized linear models.
\emph{Biometrika}, 73(1):13 to 22, 1986.

\bibitem{little1961}
J. D. C. Little.
A proof for the queuing formula: $L = \lambda W$.
\emph{Operations Research}, 9(3):383 to 387, 1961.

\bibitem{miller2024errorbars}
E. Miller.
Adding error bars to evals: A statistical approach to language model evaluations.
\emph{arXiv preprint arXiv:2411.00640}, 2024.

\bibitem{newman2023}
M. E. J. Newman.
Efficient computation of rankings from pairwise comparisons.
\emph{Journal of Machine Learning Research}, 24(238):1 to 25, 2023.

\bibitem{rao1967}
P. V. Rao and L. L. Kupper.
Ties in paired-comparison experiments: A generalization of the Bradley-Terry model.
\emph{Journal of the American Statistical Association}, 62(317):194 to 204, 1967.

\bibitem{rawles2023aitw}
C. Rawles, A. Li, D. Rodriguez, O. Riva, and T. Lillicrap.
Android in the Wild: A large-scale dataset for Android device control.
In \emph{Advances in Neural Information Processing Systems 36 (NeurIPS), Datasets and Benchmarks Track}, 2023.

\bibitem{sainz2023contamination}
O. Sainz, J. A. Campos, I. Garc\'{\i}a-Ferrero, J. Etxaniz, O. Lopez de Lacalle, and E. Agirre.
NLP evaluation in trouble: On the need to measure LLM data contamination for each benchmark.
In \emph{Findings of the Association for Computational Linguistics: EMNLP 2023}, 2023.

\bibitem{white1980}
H. White.
A heteroskedasticity-consistent covariance matrix estimator and a direct test for heteroskedasticity.
\emph{Econometrica}, 48(4):817 to 838, 1980.

\bibitem{white2024livebench}
C. White, S. Dooley, M. Roberts, A. Pal, B. Feuer, S. Jain, R. Shwartz-Ziv, N. Jain, K. Saifullah, S. Dey, Shubh-Agrawal, S. S. Sandha, S. Naidu, C. Hegde, Y. LeCun, T. Goldstein, W. Neiswanger, and M. Goldblum.
LiveBench: A challenging, contamination-limited LLM benchmark.
In \emph{Proceedings of the 13th International Conference on Learning Representations (ICLR)}, 2025.

\bibitem{xie2024osworld}
T. Xie, D. Zhang, J. Chen, X. Li, S. Zhao, R. Cao, T. J. Hua, Z. Cheng, D. Shin, F. Lei, Y. Liu, Y. Xu, S. Zhou, S. Savarese, C. Xiong, V. Zhong, and T. Yu.
OSWorld: Benchmarking multimodal agents for open-ended tasks in real computer environments.
In \emph{Advances in Neural Information Processing Systems 37 (NeurIPS), Datasets and Benchmarks Track}, 2024.

\bibitem{zermelo1929}
E. Zermelo.
Die Berechnung der Turnier-Ergebnisse als ein Maximumproblem der Wahrscheinlichkeitsrechnung.
\emph{Mathematische Zeitschrift}, 29:436 to 460, 1929.

\bibitem{zheng2023judging}
L. Zheng, W.-L. Chiang, Y. Sheng, S. Zhuang, Z. Wu, Y. Zhuang, Z. Lin, Z. Li, D. Li, E. P. Xing, H. Zhang, J. E. Gonzalez, and I. Stoica.
Judging LLM-as-a-judge with MT-Bench and Chatbot Arena.
In \emph{Advances in Neural Information Processing Systems 36 (NeurIPS), Datasets and Benchmarks Track}, 2023.

\bibitem{zhou2024webarena}
S. Zhou, F. F. Xu, H. Zhu, X. Zhou, R. Lo, A. Sridhar, X. Cheng, T. Ou, Y. Bisk, D. Fried, U. Alon, and G. Neubig.
WebArena: A realistic web environment for building autonomous agents.
In \emph{Proceedings of the 12th International Conference on Learning Representations (ICLR)}, 2024.

\end{thebibliography}
\end{document}